\documentclass[lettersize,journal]{IEEEtran}
\usepackage{amsmath,amsfonts}
\usepackage{algorithmic}
\usepackage{algorithm}
\usepackage{array}
\usepackage[caption=false,font=normalsize,labelfont=sf,textfont=sf]{subfig}
\usepackage{textcomp}
\usepackage{stfloats}
\usepackage{url}
\usepackage{verbatim}
\usepackage{graphicx}
\usepackage{cite}

\usepackage{subcaption}

\usepackage{enumitem}
\usepackage{booktabs}
\usepackage{multirow}
\usepackage{threeparttable}
\usepackage{xcolor} 

\usepackage{amssymb}   
\usepackage{bbding}  

\begin{document}

\title{CORE Planner: Contextual-memory Oriented Reinforcement-learning in Unknown Environments for Robot Navigation%
\thanks{%
\textcopyright~2026 IEEE. Personal use of this material is permitted.
Permission from IEEE must be obtained for all other uses, in any current or
future media, including reprinting/republishing this material for advertising
or promotional purposes, creating new collective works, for resale or
redistribution to servers or lists, or reuse of any copyrighted component of
this work in other works.}}

\author{
~Jintao~Kong$^{1}$,
~Zhihao~Zhang$^{1}$,
~Weihuang Chen$^{1*}$,
~Liming Chen$^{1}$,
~Zhongyu Guo$^{1}$,
~Shuaiyu Liu$^{1}$,
and ~Hongbin Sun$^{1}$,~\IEEEmembership{Senior Member,~IEEE}
\thanks{This research was supported by the National Natural Science Foundation of China (No.62503381, No. U24A20291).}
\thanks{$^{1}$State Key Laboratory of Human-Machine Hybrid Augmented Intelligence, Institute of Artificial Intelligence and Robotics, Xi'an Jiaotong University}
\thanks{*Corresponding Author: Weihuang Chen, chenwh@xjtu.edu.cn.}
}

\markboth{Journal of \LaTeX\ Class Files,~Vol.~14, No.~8, August~2021}%
{Shell \MakeLowercase{\textit{et al.}}: A Sample Article Using IEEEtran.cls for IEEE Journals}

\IEEEpubid{0000--0000/00\$00.00~\copyright~2026 IEEE}

\maketitle

\begin{abstract}
Autonomous navigation in unknown environments requires a robot to efficiently reach a predefined goal while exploring without prior maps. Although progress has been made in this area, most existing works still rely on traditional planning methods with hand-crafted rules, while learning-based methods often suffer from limited environmental memory and challenges in simulation-to-real (sim-to-real) transfer. To overcome these limitations, we propose a Contextual-memory Oriented Reinforcement-learning (CORE) planner for robot navigation in unknown environments. The proposed CORE planner effectively combines the core advantages of traditional and learning-based methods. Specifically, our method uses a sparse visibility graph for structured environment representation, reducing the computational overhead of dense grid maps, and employs a Transformer network to achieve a holistic environmental understanding, thereby significantly improving navigation efficiency. Moreover, we introduce a visibility graph-based graph sparsification method and a contextual memory mechanism, which alleviates local optima and enhances computational performance in large-scale scenes. Finally, our approach achieves zero-shot sim-to-real transfer after training solely on image-based environments, requiring no fine-tuning. Experimental results show that CORE Planner consistently outperforms state-of-the-art methods, including the traditional FAR Planner and all learning-based baselines, across representative environments, reducing travel distance by 13\% over traditional FAR Planner and by up to 48\% relative to learning-based baselines, with larger gains observed in more complex environments. In real-world scenarios, CORE successfully navigates without human intervention, showcasing zero-shot sim-to-real transfer. Code
is available at \url{https://github.com/BBD00/core_planner}.
\end{abstract}


\section{Introduction}
\begin{figure}[!h]
    \centering
    \includegraphics[width=0.9\linewidth]{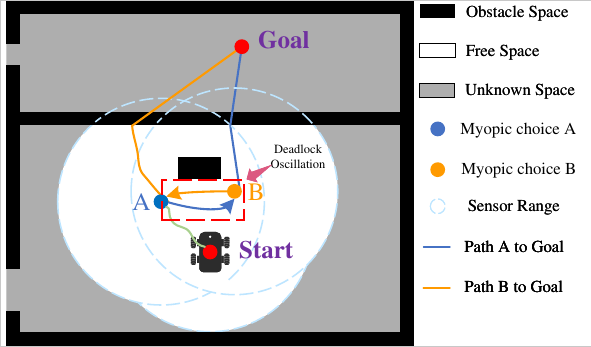}
    \caption{Illustration of the unknown-environment navigation problem. The robot incrementally explores an initially unknown environment using onboard sensing, where the white region denotes the currently observed free space, the gray region denotes unknown space, and black regions denote obstacles. At each step, the robot updates its environmental belief within the sensor range and selects the next waypoint toward the goal. Positions A and B illustrate two candidate local decisions: a myopic choice near obstacles may lead to oscillatory behavior and inefficient progress, whereas a context-aware choice helps the robot escape the local deadlock and continue toward the goal.}
    \label{fig:description}
\vspace{-12pt}
\end{figure}

\IEEEPARstart{I}{n} recent years, robotics technology has made remarkable progress, with autonomous navigation demonstrating significant potential in critical applications such as post-disaster search and rescue, and planetary exploration \cite{nahavandi2025comprehensive, al2024advancements}. These scenarios often require robots to navigate toward target locations in unknown environments without prior maps, as illustrated in Fig.~\ref{fig:description}. Throughout the navigation process, the robot continuously acquires sensor data, updates its environmental representation, and decides on subsequent waypoints to ultimately reach the goal. In other words, the robot must minimize unnecessary exploration while achieving its goal, balancing efficiency and reliability under conditions of uncertainty.

Research on autonomous navigation in unknown environments has long been dominated by traditional methods. These approaches typically incrementally construct an environmental map and employ heuristic functions to select the next target point. However, they heavily rely on predefined rules and idealized environmental assumptions, often leading to local optima, such as dead ends, and eventual task failure\cite{wang2024online}. Furthermore, their widespread dependence on dense grid-based representations results in low computational efficiency, especially in large-scale settings. Given the inefficiency of dense grid-based representations in large-scale environments, subsequent research has explored sparse alternatives. Among them, FAR Planner \cite{far} employs a visibility graph-based representation to substantially improve computational performance.
\IEEEpubidadjcol

With advances in deep learning and reinforcement learning (RL), some researchers have begun to explore learning-based solutions for path planning in unknown environments. For instance, NavRL \cite{navrl} and YOPO \cite{yopo} enabled drone path planning in simple unknown environments. However, these methods assume cylindrical or forest-like obstacle structures, which limits their applicability in complex indoor scenes. Methods such as GDAE \cite{goal} rely on classical planning methods (e.g., POI \cite{ou2024poi}, Voronoi \cite{voronoi}) to output intermediate targets for an RL-based local planner, thereby accomplishing navigation tasks. While these hybrid approaches have spurred the development of numerous myopic local planners based on RL \cite{ou2024poi,voronoi,huang2023goal}, we argue that such workarounds still depend on handcrafted rules and lack holistic environmental understanding.

Naturally, we aim to integrate the sparse representation advantage of traditional visibility graphs with the complex rule-learning capability of learning-based methods. In parallel, large language models (LLMs) \cite{attention,brown2020language} have recently shown remarkable progress, especially in long-context memory. This capability motivates us to integrate such contextual memory into robotic navigation. To this end, we propose CORE planner, a \textbf{C}ontextual-memory \textbf{O}riented \textbf{RE}inforcement-learning method for robot navigation in unknown environments. As the first navigation method to combine visibility graphs with contextual memory, it effectively integrates the advantages of traditional planning, learning-based strategies, and large-scale memory. This integration is accomplished by introducing novel environmental representations and contextual memory mechanisms, rather than modifying the network architecture itself. Specifically, we abstract the environment by transforming dense representations based on grids into a sparse visibility graph. This topological abstraction retains only essential geometric connectivity while discarding redundant spatial details, thereby significantly enhancing computational efficiency. The visibility graph representation is then fed into a Transformer-based encoder-decoder network. The encoder extracts structural features from the visibility graph while integrating contextual historical actions to achieve a more comprehensive global understanding. The decoder takes the current robot position node as the query and the surrounding neighbor nodes as keys and values, outputting a probability distribution over candidate next waypoints. 

The main contributions of this work are summarized as follows:

\begin{itemize}
\item To the best of our knowledge, this is the first reinforcement learning-based navigation method that achieves zero-shot sim-to-real transfer in unknown environments. 

\item  We propose a visibility graph-based reinforcement learning method that enables contextual environment memory for robotic navigation, effectively addressing the challenge of local optima.

\item We introduce a dynamic graph sparsification method built upon the visibility graph, which improves decision-making efficiency in large-scale environments.

\item Extensive experiments in both simulated and physical environments demonstrate the method's robust navigation performance, real-time capability, and successful sim-to-real zero-shot transfer.
\end{itemize}
The rest of this paper is organized as follows. Section \ref{related_work} reviews related work. Section \ref{problem_formulation} formulates the problem and defines the task. Section \ref{method} presents our methodology, including the overall framework and training details. Section \ref{exp} presents comprehensive experiments including simulations, real-world tests, and ablation studies, to fully evaluate the proposed approach. Finally, Section \ref{conclusion} concludes the paper.

\begin{table}[t]
\centering
\caption{Comparison of navigation methods.}
\label{tab:sota_comparison}
\renewcommand{\arraystretch}{1.1}
\setlength{\tabcolsep}{2.5pt}
\footnotesize
\begin{tabular}{l c l c c}
\hline
\textbf{Method} & \textbf{Input} & \textbf{Rep.} & \textbf{Mem.} & \textbf{Edge} \\
\hline
FAR~\cite{far} & LiDAR & Vis. graph & No & \checkmark \\
GDAE~\cite{goal} & RGB-D & - & No & \checkmark \\
NavRL~\cite{navrl} & RGB-D & - & No & \checkmark \\
YOPO~\cite{yopo} & RGB-D & - & No & \checkmark \\
CTSAC~\cite{ctsac} & LiDAR & Hist. buffer & Short & \checkmark \\
CADRL~\cite{cadrl} & LiDAR & Sparse grid & Short & \checkmark \\
VLM/LLM~\cite{yang20253d, zhu2025move} & RGB-D & Grid/Img. & Short & \XSolidBrush \\
\textbf{CORE (Ours)} & L/RGB-D & Sparse Vis. & \textbf{Full} & \checkmark \\
\hline

\multicolumn{5}{l}{\scriptsize \textit{Rep.}: Representation; \textit{Vis.}: Visibility; \textit{Mem.}: Memory; \textit{L}: LiDAR.} \\
\end{tabular}
\vspace{-12pt}
\end{table}

\vspace{-10pt}
\section{Related Work} \label{related_work}
Autonomous navigation in unknown environments has achieved considerable progress. This section reviews traditional, learning-based, and emerging VLM/LLM-driven planners for unknown environments.

\vspace{-12pt}
\subsection{Traditional Planners}

Traditional planners include search-based methods, such as Dijkstra \cite{dijkstra}, A* \cite{a_star}, D* \cite{D_star}, LPA* \cite{lpa_star}, and D* Lite \cite{D_star_lite}, and sampling-based methods, such as RRT* \cite{rrt_star}, RRT-Connect \cite{rrt_connect}, and BIT* \cite{bit_star}. Search-based methods incur increasing map-maintenance and global-search overhead as exploration expands \cite{wang2022chase}. To mitigate this, Yang et al. \cite{far} proposed FAR Planner, which uses a two-layer incremental visibility graph to reduce costs. On the other hand, sampling-based methods are dominated by RRT \cite{rrt} variants such as RRT* \cite{rrt_star}, RRT-Connect \cite{rrt_connect}, and BIT* \cite{bit_star}. While effective in high-dimensional spaces, they often require extensive planning time to locate feasible paths without prior maps and remain highly susceptible to local optima.

Consequently, the heavy reliance of traditional methods on handcrafted rules often results in becoming trapped in local minima or excessive costs for maintaining global structures. While subsequent research on visibility graph-based representations (e.g., E-Planner \cite{zhang2024planner}, FPS \cite{li2022fps}) has improved efficiency, these frameworks are still constrained by predefined features and lack holistic environmental understanding.

\vspace{-12pt}
\subsection{Learning-based Planners}
The integration of deep learning and reinforcement learning (RL) has catalyzed new paradigms for path planning, typically categorized into global and local planners. Significant progress has been made in local planning, as it avoids the need for comprehensive environmental perception. For instance, NavRL \cite{navrl} constructs cylindrical obstacle profiles using object detection and laser point clouds to train RL policies, while YOPO \cite{yopo} utilizes Euclidean Signed Distance Field (ESDF) maps to enable rapid training via direct gradient backpropagation. However, these methods often rely on predefined obstacle templates or specific map structures, limiting their applicability in complex indoor or multi-room scenarios. In the realm of global planning, algorithms such as GDAE \cite{goal} enable autonomous navigation in unknown environments by selecting next sub-goals and feeding them into a network-based local planner. Representative methods like GDAE fundamentally rely on classical global planners to generate intermediate targets for the RL-based local planner \cite{ou2024poi, voronoi}, thereby accomplishing the navigation task. These approaches have spurred the development of numerous short-horizon local planners based on reinforcement learning \cite{huang2023goal}.

However, most learning-based methods struggle with global planning due to limited environmental memory and fixed-size neural inputs.
Conventional CNN architectures require fixed input sizes, which is infeasible for navigation tasks in unknown environments where the explored area continuously expands, leading to dynamic input dimensions \cite{cong2023review}. Consequently, most current learning-based methods only process current sensor readings or use predefined input scales. 

\vspace{-12pt}
\subsection{VLM/LLM-based Planners}

Recent advances in large vision-language models (VLMs) and large language models (LLMs) have inspired navigation frameworks that leverage semantic reasoning or natural-language instructions for decision making \cite{yang20253d, zhu2025move, schumann2024velma}. These approaches typically formulate navigation as a vision-language grounding problem, where agents interpret textual goals and generate corresponding actions. Although effective for instruction-following tasks, they differ from navigation in unknown environments in both formulation and system design. In particular, VLM/LLM-based methods often rely on large foundation models \cite{bai2023qwen}, making real-time deployment on edge platforms such as Jetson Orin NX challenging. Moreover, they emphasize semantic understanding rather than geometric reasoning \cite{ren2024explore}, and therefore do not explicitly address structured environment representation, partial observability, or local optima caused by inconsistent belief updates \cite{wang2025expand}.

In contrast, our method is designed as a geometry-driven and edge-efficient planner tailored for unknown environments. It performs high-frequency waypoint selection using structured spatial cues rather than semantic inputs.
Building on this perspective, we propose CORE planner, a Contextual-memory Oriented Reinforcement-learning method for robot navigation in unknown environments. This framework models the environment through a visibility graph and employs a Transformer-based encoder–decoder architecture, with no fixed input size constraints imposed.
This design enables efficient decision-making in large-scale environments. The resulting policy further exhibits strong zero-shot sim-to-real transfer, as it does not require any additional real-world fine-tuning.

\vspace{-6pt}
\section{Problem Formulation} \label{problem_formulation}
In this section, we formalize the navigation problem in unknown environments and highlight the limitations of classical greedy formulations under partial observability. We first describe the task setting and belief update process, followed by the classical next waypoint selection formulation and its practical challenge.

\vspace{-12pt}
\subsection{Task Description}
The core objective of unknown environment navigation is to enable a robot to reach a predefined goal autonomously without prior environmental maps, while continuously updating environmental awareness via sensor measurements \cite{far}. The task environment is modeled as a 2D occupancy grid map \(E\), consisting of free space \(E_f\) and obstacle space \(E_o\). The robot’s state \(S\) includes its position and other relevant variables. Its belief about the environment, denoted \(B\), comprises believed obstacle space \(B_o\), unknown space \(B_u\), and free space \(B_f\). At each step, the robot updates its belief using sensor measurements \(M\) such that \(B = B \cup M\), until the goal is reached. The objective is to find the shortest collision-free path \(p^*\) to the goal by selecting, at each time step, the optimal next waypoint or local trajectory according to \cite{cadrl}:
\begin{align}
\label{f1}
p^*(t)=\arg\min_{p\in\mathcal{P}(t)}\; \mathrm{Cost}\!\big(p;\; S(t-1), B(t-1)\big),
\end{align}
where $\mathcal{P}(t)$ denotes the set of admissible candidate local plans or next waypoints at time $t$, and \(\text{Cost}\) is an objective function usually incorporating metrics such as distance to goal, heading deviation, path smoothness, etc.

\vspace{-12pt}
\subsection{Practical Challenges} 
Although classical formulations can theoretically incorporate historical actions \cite{lpa_star}, in practical unknown environment navigation the cost is almost always computed only from the current belief map, which lacks explicit encoding of the robot’s past motion. Under partial observability \cite{far, ctsac}, this often leads to inconsistent evaluations across consecutive steps, causing oscillatory behavior or entrapment in local optima. As illustrated in Fig. \ref{fig:description}, in a room-to-room scenario, the robot may greedily choose the shortest path, which erroneously passes through a wall and is therefore infeasible. This leads to a local optimum where the robot becomes trapped. Due to partial observability, the robot remains unaware of a door behind it and becomes trapped, exhibiting oscillatory navigation behavior. Furthermore, sensor noise can create false beliefs that certain blocked regions are traversable, reinforcing the robot’s misinterpretation. In this scenario, the robot initially selects the optimal path point $A$ based on its current beliefs. However, as it moves toward $A$, new sensor data is received. This updated information suddenly makes a previous waypoint, $B$, appear optimal again. Consequently, the robot reverses its direction and moves back toward $B$. This cycle repeats indefinitely, a phenomenon we term \textbf{Navigation Deadlock Oscillation (NDO)}, analogous to a multithreaded deadlock in computer programs. This phenomenon manifests as a characteristic back-and-forth movement pattern. The quantitative evidence for this behavior is presented in Experiment \ref{exp3} and illustrated in Fig. \ref{fig:description}. 

The proposed CORE planner introduces contextual action memory to effectively mitigate this fundamental issue, offering a novel solution for robust long-horizon navigation under uncertainty. Therefore, in our proposed framework, the objective function is extended from Eq.\ref{f1} to the following form:
\begin{equation}
p^*(t)=\arg\min_{p\in\mathcal{P}(t)}\; \mathrm{Cost}\!\big(p;\; S(0),\ldots,S(t-1), B(t-1)\big),
\end{equation}
where $\mathcal{P}(t)$ denotes the set of admissible candidate local plans or next waypoints at time $t$. Unlike a fixed-length history window, the term $S(0),\ldots,S(t-1)$ explicitly represents the complete historical states utilized by the planner. In the CORE planner, historical information is compactly encoded into node-wise contextual memory rather than processed as an ever-growing temporal sequence, enabling long-term history utilization without increasing inference latency as the trajectory length grows.

\vspace{-12pt}
\section{Method} \label{method}
In this section, we first formulate robot navigation in unknown environments as a sequential decision-making reinforcement learning problem and introduce the model’s reward and action design. We then detail the state representation with contextual action memory and the visibility graph sparsification algorithm for large-scale environments. Subsequently, we introduce our contextual-memory network architecture, followed by details of the model training procedure.

\vspace{-12pt}
\subsection{Navigation in Unknown Environments as a RL Problem}
The navigation problem in unknown environments is formulated as a Markov Decision Process (MDP), denoted by the tuple \{$S$, $A$, $P$, $R$, $\gamma$\}, where $S$ represents the state space, $A$ the action space, $P$ the state transition probability, $R$ the reward function, and $\gamma < 1$ the discount factor for future rewards. To achieve efficient navigation in unknown environments, the objective of the reinforcement learning agent is to learn an optimal policy $\pi^{*}(a_t|s_t)$ that maximizes the expected cumulative return:
\begin{align}
\pi^* = \arg\max_{\pi} 
\mathbb{E}_{\pi}\!\left[
\sum_{\tau=0}^{T_{\mathrm{ep}}} \gamma^{\tau} R(s_{\tau}, a_{\tau})
\right].
\end{align}

\textbf{Action $A$:} Given that the robot's position node and its collision-free edges (visible edges) to neighboring nodes have been constructed, the agent's action is naturally defined as selecting one of the nodes connected by these edges as its next waypoint. It is noteworthy that this selection is implemented via a Pointer Network \cite{vinyals2015pointer}, which enables flexible input and output dimensions without predefined sizes, thereby ensuring adaptability across environments.

\textbf{Reward $R$:} Our reward consists of five components:
\begin{align}
r = \lambda_{1}r_{\mathrm{goal}} 
+ \lambda_{2}r_{\mathrm{frontier}} 
+ \lambda_{3}r_{\mathrm{ds}} 
+ \lambda_4r_{\mathrm{stay}} 
+ \lambda_5r_{\mathrm{f}}.
\end{align}
where each term represents a specific reward type weighted by its corresponding coefficient $\lambda_i$. The design and role of each reward will be elaborated in the following paragraphs.

\subsubsection{Goal Distance Reward \( r_{\mathrm{goal}} \)} This reward encourages the agent to move toward the goal. It is defined as:  
\begin{align}
r_{\mathrm{goal}} = \frac{1}{\|r_t-\mathrm{goal}\|},
\end{align}
where $\mathrm{goal}$ denotes the goal position and $r_t$ represents the robot's location at time $t$.
\subsubsection{Frontier Reward $r_\mathrm{frontier}$} To promote exploration in the early stages, this reward incentivizes the agent based on the number of newly observed frontier points, written as:  
\begin{align}
r_{\mathrm{frontier}} = F_t - F_{t-1},
\end{align}
where $F_t$ indicates the number of frontier points at time $t$.
\subsubsection{Position difference reward $r_{\mathrm{ds}}$} This item penalizes stagnation by calculating the Euclidean distance between the robot's current position and its previous position, expressed as:
\begin{align}
r_{\mathrm{ds}} = -\| r_t - r_{t-1} \|.
\end{align}
\subsubsection{Stagnation Penalty}
To further discourage the agent from being trapped in local optima, an additional penalty is applied based on consecutive time steps without movement:  
\begin{align}
    r_{\mathrm{stay}} =
    \begin{cases}
    \mathrm{stay} - 5, & \text{if } \mathrm{stay} > 5,\\
    0, & \text{otherwise}.
    \end{cases}
\end{align}
where $\mathrm{stay}$ represents the duration of continuous stagnation.

\subsubsection{Finish reward $r_\mathrm{f}$}  
A finish reward is provided when the agent successfully reaches the goal, thereby terminating the episode. It is defined as:  
\begin{align}
    r_{\mathrm{f}} =
    \begin{cases}
    20, & \text{if } \| r_t - \mathrm{goal} \| < \epsilon,\\
    0, & \text{otherwise}.
    \end{cases}
\end{align}
where $\epsilon$ is a distance threshold.

\begin{figure}[!h]
    \centering
    \subfloat[]{
        \includegraphics[width=0.45\linewidth]{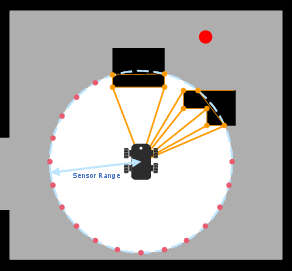}
        \label{fig:part_known}
    }
    \subfloat[]{
        \includegraphics[width=0.45\linewidth]{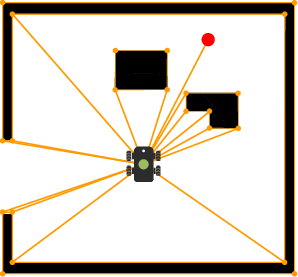}
        \label{fig:all_known}
    }
    \caption{
        Illustration of visibility graph construction in partially-known and fully-known environments.
        (a) Partially-known; (b) Fully-known.
        Blue dashed circles denote LiDAR ranges, orange points and segments indicate visibility vertices and edges, magenta points mark boundary points, and red points denote targets.
    }
    \label{fig:graph_show}
    \vspace{-12pt}
\end{figure}

\subsection{State with Contextual Action Memory}
At time step $t$, the state of the agent is represented by a visibility graph structure $G_t = (N_t, E_t)$. The nodes $N_t$ and edges $E_t$ of this graph are constructed by converting sensor data into a visibility graph, as illustrated in Fig. \ref{fig:graph_show}. The edges $E_t$ correspond to collision-free connections between nodes in the visibility graph. Each node $n_i \in N_t$ consists of four components: its spatial coordinates $p_i$, a utility value $u_i$, the distance to the goal $d_i$, and a trajectory indicator $v_i$. The utility value $u_i$ is defined as the number of observable frontier points around the node. Here, frontier points represent the boundary between known and unknown regions, suggesting areas that are likely to lead toward the goal. The goal distance $d_i$ represents the Euclidean distance from the current node to the target. The trajectory indicator $v_i$ records the number of times the agent has visited the current node $n_i$ since the start of the task, thereby characterizing the agent’s contextual action memory. Its computation is given as follows:
\begin{equation}
\label{eq:trajectory_indicator}
v_i = \sum_{\tau=1}^{t}
\mathbb{I}\!\left(D(r_\tau, p_i) < \epsilon\right),
\end{equation}
where $D(\cdot,\cdot)$ represents the Euclidean distance between two spatial coordinates, $\mathbb{I}(\cdot)$ is the indicator function (i.e., \(\mathbb I(\text{True}) = 1\) and \(\mathbb I(\text{False}) = 0\)), $r_\tau$ is the agent's position at time $\tau$, $p_i$ is the spatial coordinate of node \(n_i\),  $\epsilon$ is a distance threshold and $t$ represents the current time step.

In contrast to methods such as CADRL \cite{cadrl}, our trajectory indicator is not limited to a binary value, enabling the agent to perceive its movement history more comprehensively and thus better escape local optima. Therefore, each node $n_i$ in the visibility graph $G_t$ can be represented as a feature vector $\mathbf{x}_i$, formed by concatenating its physical-spatial and internal-state components:
\begin{equation}
\label{vector_node}
\mathbf{x}_i = [p_i \oplus u_i \oplus d_i \oplus v_i]
\end{equation}
where $\oplus$ denotes the concatenation operation, and $v_i$ is the contextual trajectory indicator explicitly defined in Eq.~\eqref{eq:trajectory_indicator}, which represents the contextual memory.

Although the visibility graph already offers a relatively sparse representation of the environment, we propose a graph sparsification method based on the visibility graph to further enhance computational efficiency in large-scale environments. The pseudocode is provided in Algorithm \ref{alg:alg1}. Specifically, we first define a non-sparsified region around the robot to preserve full environmental details in its immediate vicinity. For areas beyond this region, we perform node fusion by clustering interconnected visible points within a predefined radius threshold. These clustered points are then merged into representative nodes located at their centroid, with all original edges inherited accordingly. This aggregation process significantly reduces the number of nodes while maintaining graph connectivity.

To provide a clear comparison between the proposed graph sparsification method and conventional non-sparsified approaches, we analyze the computational complexity to demonstrate the scalability of our framework. Let $N$ and $N'$ denote the number of raw and sparsified nodes, respectively. Using a spatial index, the graph sparsification process incurs a cost of $T_{\mathrm{sparsify}} \approx O(N \log N)$. The subsequent Transformer encoding requires $T_{\mathrm{transformer}} = O(N'^2)$, resulting in a total complexity of $T_{\mathrm{total}} = O(N \log N + N'^2)$. In contrast, a non-sparsified approach scales quadratically with the raw input size, i.e., $T_{\mathrm{raw}} = O(N^2)$. Given that the sparsification ratio $\alpha = N'/N$ is typically small (empirically $\alpha \leq 0.5$), our method significantly reduces computational overhead. This theoretical analysis is fully consistent with the real-time performance observed in our simulation (Table~\ref{table:comparison_img}) and physical experiments on Jetson Orin NX (Table~\ref{table:comparison_exp3}).

\begin{algorithm}[!h]
\caption{Graph Sparsification of Visibility Graph.}\label{alg:alg1}
\begin{algorithmic}[1]
\STATE \textbf{Input:} Visibility Graph $G = (N, E)$
\STATE \textbf{Output:} Sparsified Visibility Graph $G' = (N', E')$
\STATE Initialize $N' \gets \emptyset$, $E' \gets \emptyset$
\STATE $N_{\text{near}} \gets \{ n \in N \mid \| \mathbf{p}_n - \mathbf{p}_r \| \leq r \}$  // Define non-sparsified region
\STATE $N' \gets N' \cup N_{\text{near}}$ // Preserve nearby nodes
\FOR{$n_i \in N \setminus N_{\text{near}}$}
    \IF{$n_i$ not clustered}
        \STATE $C \gets \{ n_j \in V \mid \| \mathbf{p}_{n_j} - \mathbf{p}_{n_i} \| \leq d_{\text{cluster}} \ \& \ (n_i, n_j) \in E \}$  // Find interconnected nodes within threshold
        \IF{$|C| > 1$ }
            \STATE $\mathbf{p}_{\text{centroid}} \gets \frac{1}{|C|} \sum_{n_j \in C} \mathbf{p}_{n_j}$ // Clustering nodes
            \STATE $n_{\text{new}} \gets (\mathbf{p}_{\text{centroid}})$ 
            \STATE $N' \gets N' \cup \{ n_{\text{new}} \}$
            \STATE $E' \gets E' \cup \{ (n_{\text{new}}, e) \mid \exists n_j \in C \text{ s.t. } (n_j, e) \in E \}$   // Inherit all edges
        \ELSIF {Single node remains}
            \STATE $N' \gets N' \cup \{ n_i \}$
        \ENDIF
    \ENDIF
\ENDFOR
\STATE \textbf{return} $G' = (N', E')$
\end{algorithmic}
\end{algorithm}
\vspace{-15pt}

\begin{figure*}[!ht]
    \centering
    \includegraphics[width=0.85\linewidth]{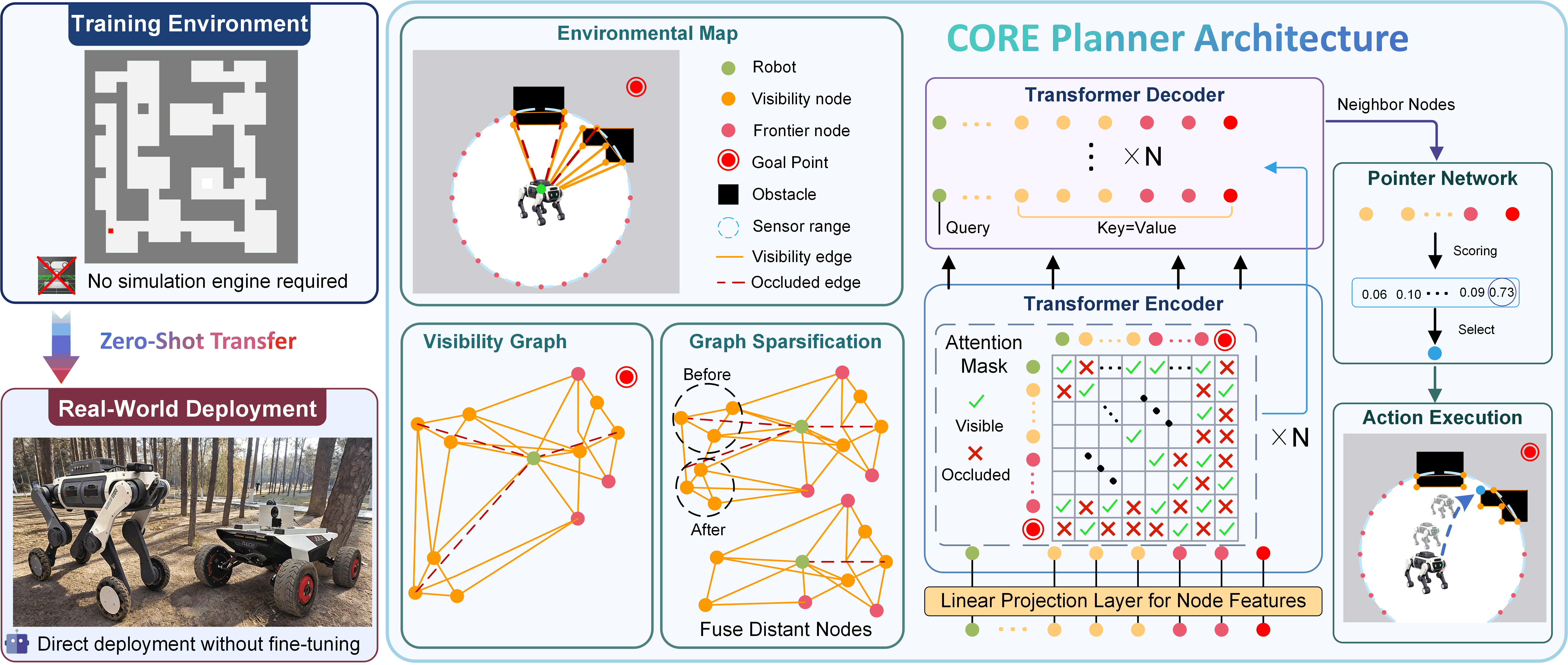}
    \caption{
    The overall pipeline of the proposed CORE planner for zero-shot navigation. A visibility graph (orange nodes and edges) is incrementally constructed from LiDAR point clouds by adding the robot (green), the goal (red), and frontier points (magenta). This graph, augmented with contextual action memory, is processed by a Transformer-based encoder–decoder to extract spatial and historical features. A pointer network then selects one of the robot’s neighboring nodes (blue) as the next waypoint. The entire model enables zero-shot transfer from simulation to real-world deployment without any fine-tuning.
    }
    \label{network}
    \vspace{-16pt}
\end{figure*}

\subsection{Visibility Graph-Based Network Architecture}

Inspired by \cite{cao2023ariadne}, we employ a standard Transformer-based encoder–decoder architecture, as illustrated in Fig.~\ref{network}. We explicitly note that this architecture is not positioned as a contribution of our work. Instead, it is adopted as a well-established and effective backbone for processing the proposed visibility graph representation together with the contextual action memory. The use of a standard backbone also offers flexibility for future extensions, where more advanced architectural designs may be incorporated to further enhance performance. Specifically, the encoder module is designed to extract global features of the known environment along with historical action features of the agent. Based on these encoded representations, the decoder infers dependencies between the current agent node and its visible edges, ultimately outputting a probability distribution over all visible edges, which serves as the action policy.

\textbf{Encoder}: The node features $\mathbf{x}_i$ from the visibility graph $G_t$ are first aggregated into a global feature matrix $\mathbf{X}$, where each row represents the complete feature vector of a node, including the contextual memory indicator $v_i$:
\begin{equation}
\mathbf{X} = \big[\mathbf{x}_1\ \mathbf{x}_2\ \dots\ \mathbf{x}_N\big]^{\top} \in \mathbb{R}^{N \times D_x}
\end{equation}
where $N$ is the total number of nodes in the current graph, and $D_x = \dim(\mathbf{x})$ is the fixed dimensionality of the input feature vector. 

These raw features are then projected into a $D_h$-dimensional latent embedding space via a learnable linear layer to form the initial input $\mathbf{H}^{(0)}$ for the Transformer encoder:
\begin{equation}
\mathbf{H}^{(0)} = \text{ReLU}(\mathbf{X}\cdot W_{emb} + \mathbf{B}_{emb}),
\end{equation}
where $\mathbf{H}^{(0)} \in \mathbb{R}^{N \times D_h}$ is the resulting $D_h$-dimensional node embedding matrix, $W_{emb} \in \mathbb{R}^{D_x \times D_h}$ is the learnable weight matrix that maps the input dimension $D_x$ to the embedding dimension $D_h$, and $\mathbf{B}_{emb} \in \mathbb{R}^{N \times D_h}$ is the bias term. The $\text{ReLU}(\cdot)$ activation function is applied element-wise.

These features $\mathbf{H}^{(0)}$ are then processed through multiple standard Transformer encoder layers \cite{attention}. During this process, an encoder mask $M$ is applied based on the graph edges to prevent each node from attending to non-adjacent nodes. The output of the encoder, denoted as $h_e$ (referred to as environment-aware features), provides the decoder with compressed spatial information and historical robot action information. This mechanism allows the model to capture global dependencies among nodes, thereby enhancing its holistic understanding of the graph-structured environment.

\textbf{Decoder}: The environment-aware feature of the current robot node $h_r$ is first retrieved. This feature, along with the global environment-aware features $h_e$, is then fed into a standard Transformer decoder layer. Here, $h_r$ serves as the query, while $h_e$ acts as both the key and the value, enabling enhanced perception and relational modeling between the robot’s current state and the global environmental context.

Subsequently, the output of this decoder layer is concatenated with the original robot feature $h_r$, and the combined representation is projected back into a $d$-dimensional space, yielding the current robot's decoded embedding $\tilde{h}_r$. Finally, the decoded embedding $\tilde{h}_r$ and the encoded embeddings of the neighbor nodes are passed into a pointer layer\cite{vinyals2015pointer}, which is implemented as an attention mechanism. The resulting attention weights $\theta$ are directly interpreted as the action probability distribution for the current node.

\vspace{-12pt}
\subsection{Theoretical Analysis of Policy Gradient Convergence}
In this subsection, we analyze the convergence properties of the CORE planner. Since CORE is trained using the Soft Actor Critic (SAC) \cite{haarnoja2018soft} algorithm, its convergence is governed by the properties of the underlying maximum entropy reinforcement learning framework. We first outline the theoretical conditions under which SAC converges, and then explain why the structural components of CORE, including the sparse visibility graph representation and the contextual memory mechanism, preserve these conditions and thereby allow CORE to inherit the convergence guarantees of SAC.

\textbf{Convergence Foundation of Maximum Entropy RL.} The soft Q-function in SAC is defined recursively by the soft Bellman equation:
\begin{equation}
Q^\pi(s,a) = r(s,a) + \gamma \mathbb{E}_{s' \sim p}\big[V^\pi(s')\big],    
\end{equation}
where the soft value function is:
\begin{equation}
V^\pi(s) = \mathbb{E}_{a\sim\pi}\big[Q^\pi(s,a) - \alpha\log\pi(a|s)\big].
\end{equation}

The corresponding soft Bellman operator $\mathcal{T}^\pi$ is:
\begin{equation}
(\mathcal{T}^\pi Q)(s,a) = r(s,a) + \gamma \mathbb{E}_{s' \sim p}\big[V^\pi(s')\big].
\end{equation}

Following standard theoretical results, $\mathcal{T}^\pi$ is a $\gamma$-contraction under the supremum norm:
\begin{equation}
\|\mathcal{T}^\pi Q_1 - \mathcal{T}^\pi Q_2\|_\infty \le \gamma \|Q_1 - Q_2\|_\infty, \quad 0 < \gamma < 1.
\end{equation}

Thus, repeated application of $\mathcal{T}^\pi$ converges to a unique fixed point $Q^\pi$. Since SAC performs stochastic approximation of this contraction through its critic update, the learned soft Q-function converges in expectation under standard Robbins–Monro conditions on learning rates.

\textbf{Convergence of the Entropy-Regularized Policy Gradient.} Given the converged critic, the SAC actor seeks to optimize the entropy-regularized objective:
\begin{equation}
\pi^\ast = \arg\max_\pi
\mathbb{E}_{s\sim\mathcal{D}, a\sim\pi}
\big[ Q^\pi(s,a) - \alpha\log\pi(a|s) \big].
\end{equation}

The corresponding policy gradient is:
\begin{equation}
\begin{aligned}
\nabla_\theta J(\pi_\theta) = \mathbb{E}_{s\sim \mathcal{D}, a\sim\pi_\theta} \bigg[ 
\nabla_\theta\log \pi_\theta(a|s) \big( Q^\pi(s,a) - {} \\
\alpha\log \pi_\theta(a|s) \big) \bigg]
\end{aligned}.
\end{equation}

Additionally, the policy network (parameterized by \(\theta\)) is smooth and Lipschitz-continuous, a standard assumption for policy gradient convergence.

Standard stochastic approximation theory guarantees convergence to a local stationary point provided that: (i) the reward and state features are bounded, and (ii) the system dynamics satisfy the Markov property.

\textbf{Compatibility of CORE with Convergence Assumptions.} We demonstrate that CORE adheres to the necessary theoretical assumptions.
\begin{itemize}
\item Boundedness of Sparse Visibility Graph Representation: Unlike unbounded raw sensor inputs, the node features in our visibility graph $x_i = (p_i, u_i, d_i, v_i)$ are structurally bounded. The spatial coordinates $p_i$ are confined to the finite map boundaries, the utility $u_i$ and goal distance $d_i$ are normalized relative to the environment scale, and the visit count $v_i$ is clipped. Thus, the feature norm is strictly bounded:
    \begin{equation}
    \|x_i\|_2 \le B, \quad \forall i.
    \end{equation}
    This ensures that the variance of the policy gradient estimator remains finite ($\mathbb{E}[\|\nabla_\theta J\|^2] < \infty$), preventing divergence during training.
\item  Preservation of Markov Property via Contextual Memory:
    A potential theoretical concern with memory-based navigation is the violation of the Markov assumption. However, in CORE, the historical visit count $v_i$ is explicitly integrated into the state vector. By defining the augmented state space $\mathcal{S}' = \mathcal{S} \times \mathcal{V}$, the transition dynamics remain fully Markovian:
    \begin{equation}
    P(s'_{t+1} \mid s'_t, a_t) = P(s_{t+1}, v_{t+1} \mid s_t, v_t, a_t).
    \end{equation}
    Since the transitions of the visit count $v_t$ are deterministic updates based on the current action, the augmented system satisfies the Markov property required for the validity of the Bellman contraction.
\end{itemize}

Consequently, CORE operates within a well-defined MDP, inheriting the convergence guarantees of the entropy-regularized framework.

\vspace{-12pt}
\subsection{Training Details}
\begin{figure}[!h]
    \centering
    \subfloat[]{
        \includegraphics[width=0.4\linewidth]{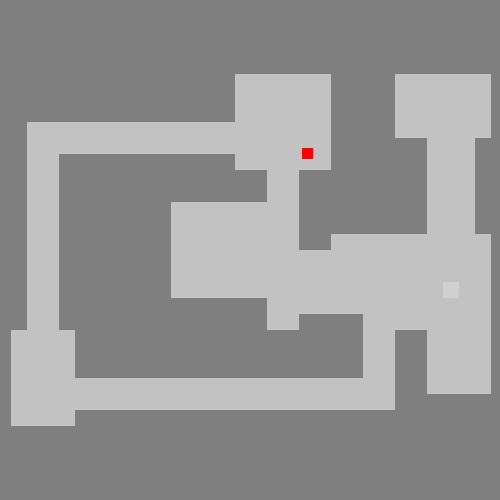}
        \label{fig:simple}
    }
    \hfill
    \subfloat[]{
        \includegraphics[width=0.4\linewidth]{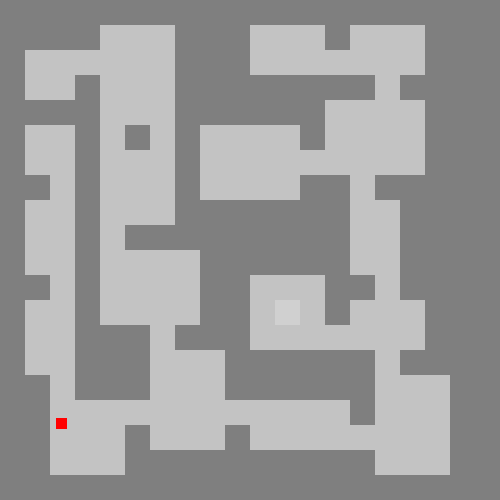}
        \label{fig:complex}
    }
    \caption{
        The training set includes both simple (a) and complex (b) environments, while the testing set consists solely of complex configurations. Dark gray areas represent non-traversable regions, light gray indicates traversable space, the white cell denotes the start point, and the red cell marks the destination.
    }
    \label{fig:env_img}
    \vspace{-8pt}
\end{figure}

Similar to CADRL, we generated both simple and highly challenging maps using the environment generator provided in \cite{chen2019self}, as shown in Fig. \ref{fig:env_img}. Each environment is represented as a $500 \times 500$ grid map, corresponding to a physical space of $100m \times 100m,$ with the sensor range set to $16m$.

For model training, we employ the SAC algorithm. The encoder and decoder described above jointly form the Actor network. The Critic network shares the same encoder and decoder architecture, with the final pointer layer replaced by a single-output linear layer. Each training episode allows a maximum of 250 steps. Both the policy and critic networks are optimized using the Adam optimizer with a learning rate of 2e-5. The experience replay buffer has a capacity of 20,000, and training proceeds with a batch size of 256. Our model was trained on a workstation equipped with an Intel Xeon E5-2660 v4 CPU and three NVIDIA GeForce RTX 2080 Ti GPUs. It is noteworthy that the training process, conducted entirely in an image-based simulated environment, achieves convergence within approximately five hours without relying on simulated physical engines such as Gazebo or Isaac Sim. This demonstrates the framework's rapid convergence capability and strong sim-to-real zero-shot transfer performance, eliminating the need for real-world fine-tuning.

\section{EXPERIMENTS} \label{exp}
To comprehensively assess the performance of the proposed CORE planner, we conducted extensive experiments in three types of environments. The experiments compare our method with SOTA traditional and learning-based approaches, and include ablation studies of different modules.

\vspace{-8pt}
\subsection{Experimental Setup and Evaluation Metrics}

\textbf{Baselines.} To ensure a fair and informative comparison, we select baselines based on their relevance to unknown-environment navigation, representativeness of different planning paradigms, and availability of reliable released implementations or model weights. \textbf{FAR Planner}~\cite{far} is chosen as the classical baseline because it is a strong visibility-graph-based planner for unknown environments. \textbf{CADRL}~\cite{cadrl} is used in the image-based experiments because it is the most directly comparable learning-based baseline in that setting: it performs learning-based navigation using a conventional grid-map representation and Transformer-based reasoning. \textbf{CTSAC}~\cite{ctsac}, \textbf{NavRL}~\cite{navrl}, and \textbf{YOPO}~\cite{yopo} are included as representative recent learning-based baselines with different sensing assumptions and planning mechanisms. Specifically, CTSAC exploits short sensor-history inputs, NavRL learns navigation policies from perception and obstacle abstractions, and YOPO leverages ESDF-based supervision for efficient policy learning.

\textbf{Experimental Setup.} Existing navigation methods are often evaluated in only a limited number of scenarios, which may overlook performance variations across environments of different complexity. To obtain a more comprehensive assessment, we evaluate the proposed method in three distinct settings. First, we test in highly challenging image-based environments, as shown in Fig.~\ref{fig:complex}, none of which are seen during training. A total of 100 such environments are regenerated according to the training settings. In this setting, CORE is compared with CADRL~\cite{cadrl}, which was also trained in similar image-based environments; for fairness and reproducibility, we directly use the official released weights and parameters of CADRL. Second, we evaluate the method in a typical autonomous exploration scenario in Gazebo~\cite{env}. Third, we conduct real-world validation to assess practical deployment capability.

\textbf{Evaluation Metrics.} We adopt the following metrics for comparison: average success rate ($S$), average travel distance ($D(m)$), average inference time ($T_e(s)$), average planning time ($T(s)$), and the number of human interventions ($I$). Here, $D(m)$, $T_e(s)$, $T(s)$, and $I$ are computed over all episodes, including failed ones. The metric $I$ records the number of times human intervention is required to guide the robot out of a local optimum. Moreover, $T_e(s)$ specifically denotes the neural network inference time, whereas $T(s)$ represents the runtime of the full planning pipeline, including sensor processing, model inference, and subsequent planning or control operations. A compact overview of the performed tests, their objectives, and the reported evaluation metrics is provided in Table~\ref{tab:exp_summary}.

\begin{table}[t]
\centering
\caption{Overview of the performed tests, their objectives, and the reported metrics}
\label{tab:exp_summary}
\footnotesize
\renewcommand{\arraystretch}{1.08}
\setlength{\tabcolsep}{2pt}
\begin{tabular}{p{2cm} p{5.3cm} p{1.0cm}}
\toprule
\multicolumn{1}{c}{\textbf{Test}} &
\multicolumn{1}{c}{\textbf{Objective}} &
\multicolumn{1}{c}{\textbf{Metrics}} \\
\midrule
Image-based & Unseen-environment generalization & $S,D,T$ \\
Gazebo comp. & Navigation, deadlock escape, efficiency & $D,I,T_e$ \\
Gazebo ablation & Memory/sparsification analysis & $D,I,T$ \\
Real-world & Deployment validation and dynamic response & $D,T,I$ \\
\bottomrule
\end{tabular}
\vspace{-10pt}
\end{table}

\vspace{-12pt}
\subsection{Comparison in Image-based Environments}

\begin{figure}[!h]
\vspace{-12pt}
    \centering
    \subfloat[]{
        \includegraphics[width=0.45\linewidth]{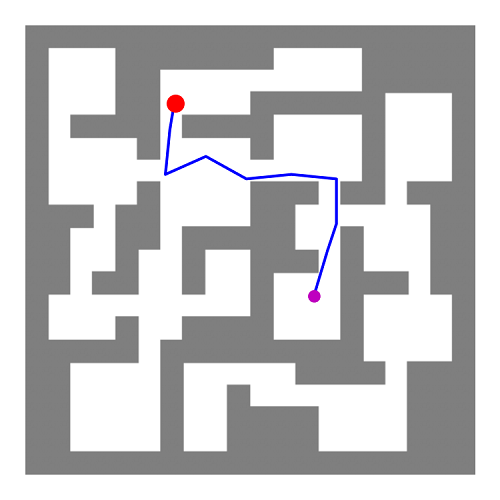}
        \label{fig:exp1_core}
    }
    \hfill
    \subfloat[]{
        \includegraphics[width=0.45\linewidth]{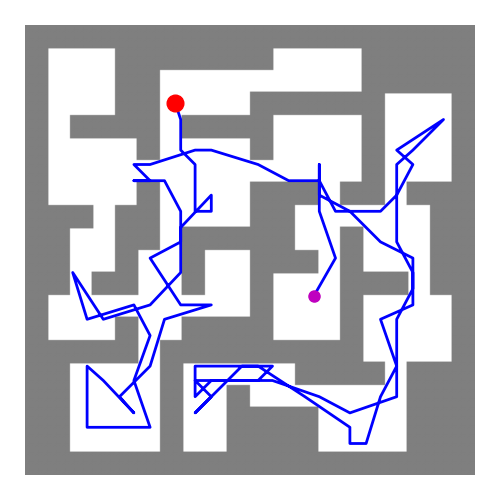}
        \label{fig:exp1_cadrl}
    }
    \caption{Comparison of representative navigation trajectories: (a) proposed CORE planner and (b) baseline CADRL.}
    \label{fig:comparison_img}
    \vspace{-15pt}
\end{figure}

We set a timeout mechanism (i.e., the maximum number of decision steps is 128) to prevent robot stagnation caused by local optimal solutions during testing; if this limit is exceeded, the test is deemed a failure. The evaluation was conducted across a total of 100 distinct image-based environments. Quantitative results are summarized in Table \ref{table:comparison_img}, and representative trajectories are illustrated in Fig. \ref{fig:comparison_img}. As shown in the table, our method achieved a 100\% average success rate, outperforming all compared algorithms. It reduced the travel distance by 20.8\% compared to CADRL. Although the planning time is slightly longer than that of CADRL, it remains within an acceptable range for a global planner.

\begin{table}[!h]
    \centering
    \caption{Comparison of properties between our method and SOTA methods}
    \label{table:comparison_img}
    \begin{tabular}{cccc}
    \toprule
    Method & $S$ &  $D(m)$ & $T(s)$ \\ 
    \midrule
    CADRL        & 97\% &391.74&\textbf{0.21}\\ 
    CORE(ours)   & \textbf{100\%} &\textbf{310.10}&0.30\\ 
    \bottomrule
    \end{tabular}
\end{table}

We note that while CADRL employs an architecture similar to CORE, it relies on a traditional grid-based representation. This results in excessively large input dimensions, which hinder the model's ability to accurately focus on relevant environmental features and ultimately lead to longer travel distances. Although CORE planner achieves a 100\% success rate in the image-based environments, we acknowledge that this outcome is partly due to the “clean” and regular obstacle geometry, which tends to produce high-quality visibility graphs. In real-world scenarios, complex structures and irregular contours may lead to suboptimal visibility graph extraction, thereby preventing the network from consistently identifying reliable structural nodes for path reasoning. Such situations may ultimately cause planning failures.

\begin{figure*}[!ht]
    \centering
    \subfloat[]{
        \includegraphics[width=0.3\linewidth]{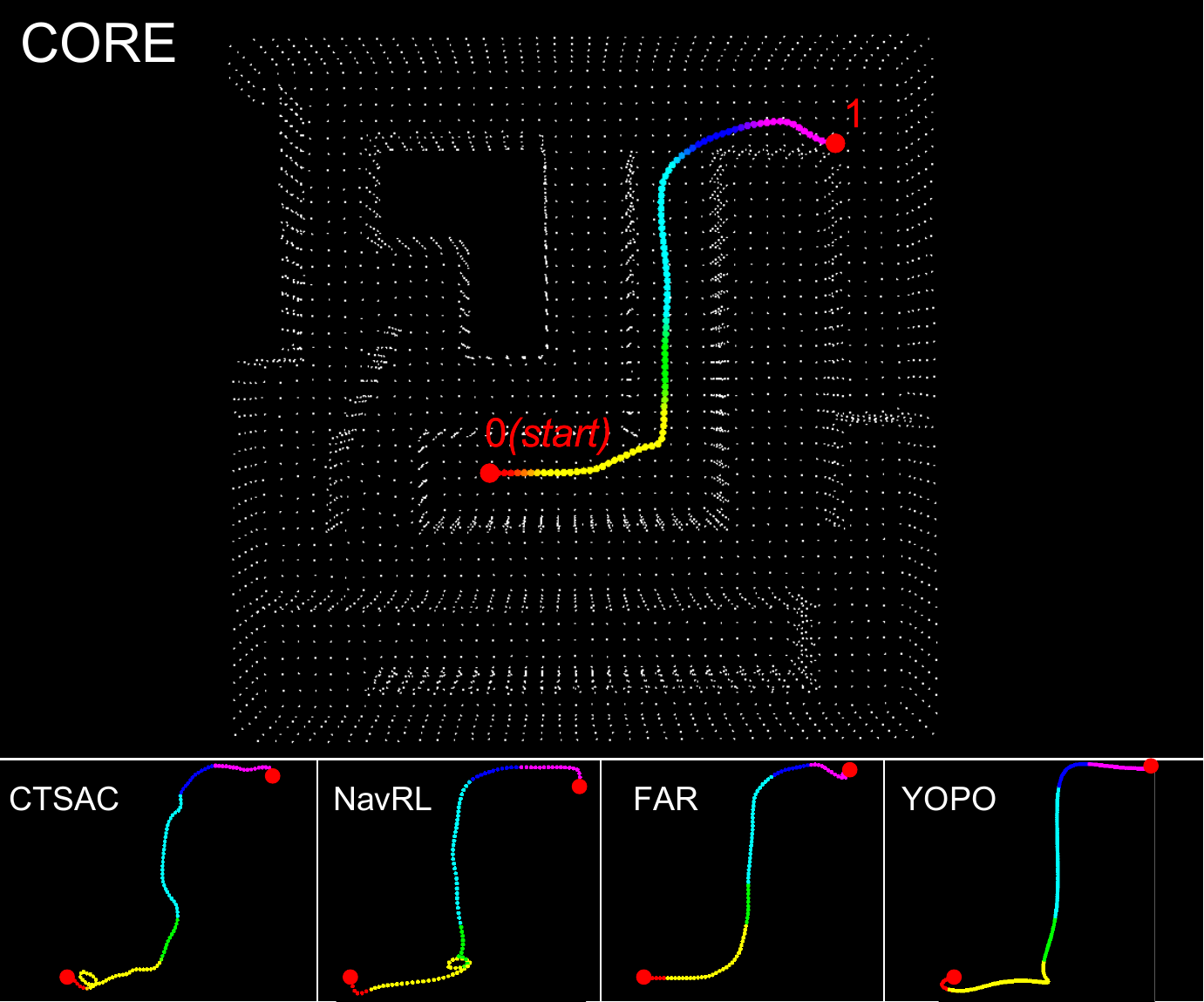}
        \label{fig:exp2_simple}
    }
    \hfill
    \subfloat[]{
        \includegraphics[width=0.3\linewidth]{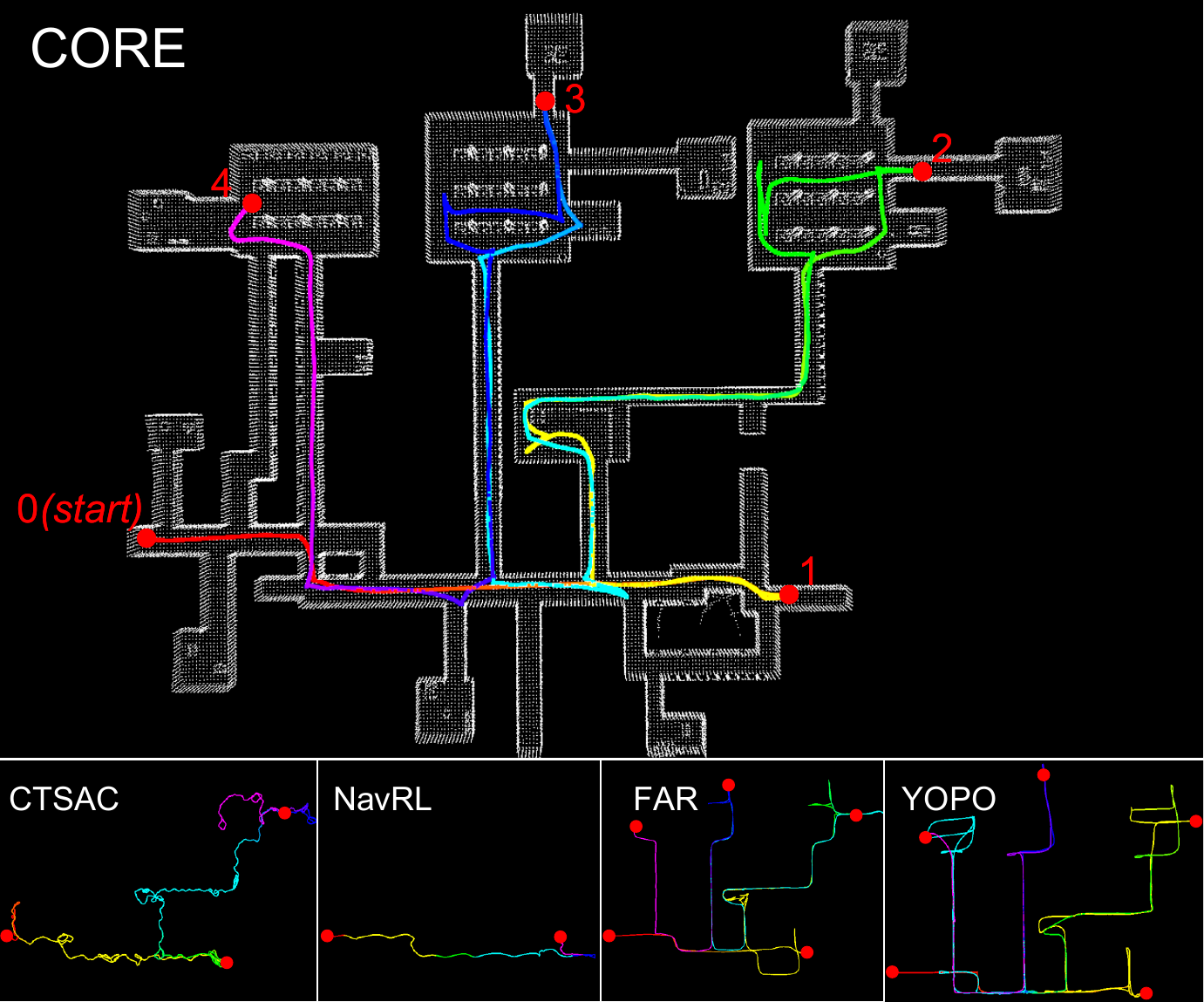}
        \label{fig:exp2_indoor}
    }
    \hfill
    \subfloat[]{
        \includegraphics[width=0.3\linewidth]{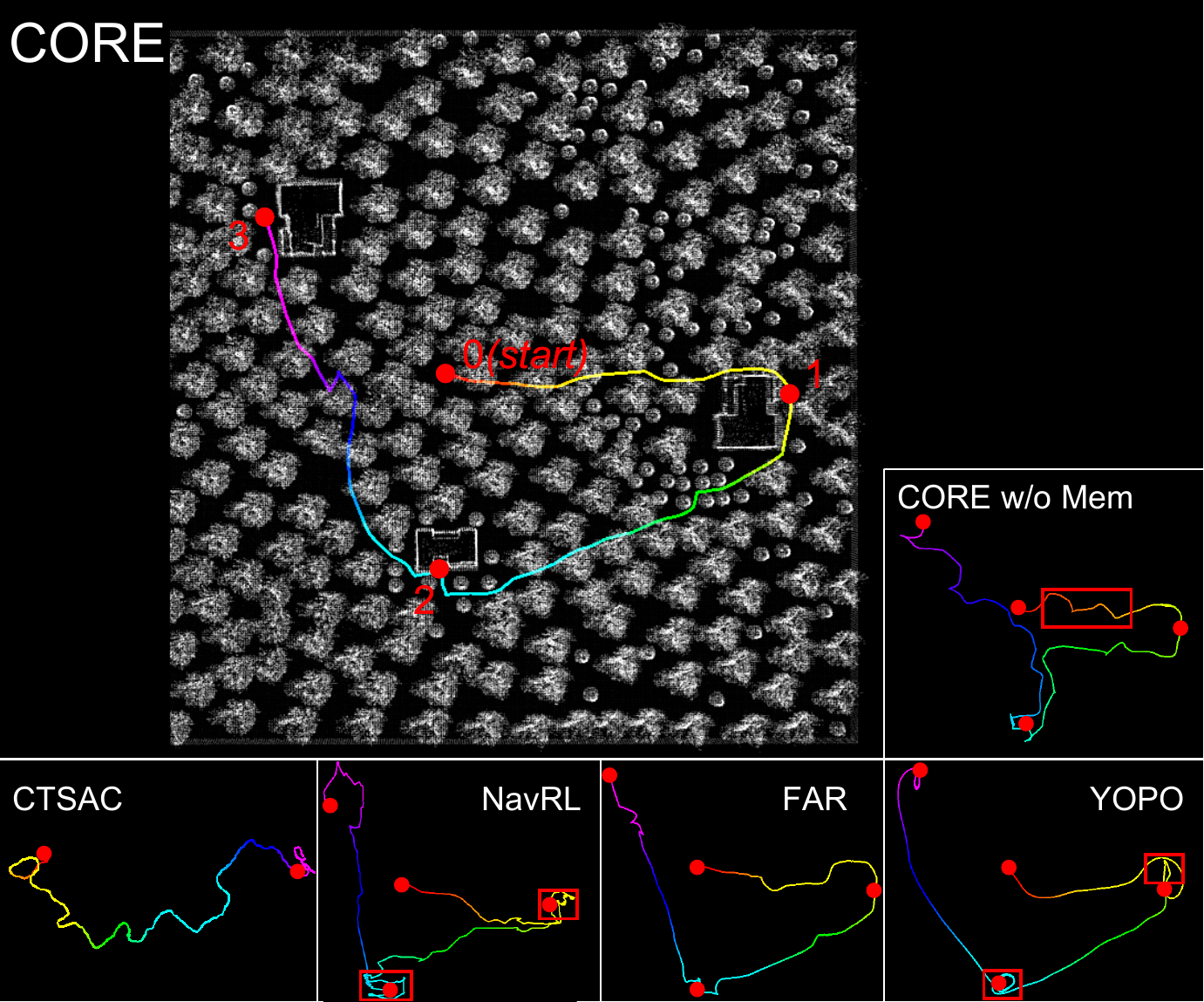}
        \label{fig:exp2_forest}
    }
    \caption{Sequential navigation through waypoints in environments of varying complexity: (a) simple environment, (b) complex indoor environment, and (c) complex forest environment. The CORE planner achieves the shortest paths compared with all other algorithms and successfully avoids local optima. 
    }
    \label{fig:comparison_exp2}
\vspace{-12pt}
\end{figure*}

\subsection{Comparison in Gazebo Environments}
The Gazebo evaluation comprises three environments: a simple indoor scenario, a complex indoor scenario, and an outdoor forest scenario, aiming to comprehensively cover potential real-world application scenarios. Furthermore, given the larger scale of the Gazebo simulation environment, we introduce a metric for the number of human interventions required to guide the robot out of local optima, enabling a more comprehensive evaluation. Navigation is considered a failure if the number of interventions exceeds 10. Experimental results are summarized in Table \ref{table:comparison_gazebo}, and representative trajectories are visualized in Fig. \ref{fig:comparison_exp2}.

Compared to the state-of-the-art traditional method FAR planner, the proposed CORE planner achieved superior performance across all three environments, reducing travel distance by up to 13\% compared to FAR Planner. Furthermore, thanks to its global contextual memory capability, CORE planner successfully completed navigation without requiring any human intervention.

Compared to learning-based approaches, CORE planner also consistently outperformed all baselines in every environment. In the simple indoor scenario, it reduced travel distance by 18.3\% compared to CTSAC, 23.3\% compared to NavRL, and 14.7\% compared to YOPO. In the complex indoor scenario, where both CTSAC and NavRL failed to complete the navigation task, CORE planner significantly reduced the travel distance by 37.1\% compared to YOPO. In the outdoor forest scenario, CTSAC again failed to navigate successfully, while CORE planner reduced travel distance by 48\% compared to NavRL, and 30.1\% compared to YOPO.

Although CTSAC retains historical sensor encodings, its training is limited to short-range navigation, rendering it ineffective in long-range tasks, a conclusion supported by the significant performance degradation observed. The variant CORE w/o Mem, while effective in short-distance navigation, exhibited extensive back-and-forth motion in long-range scenarios. This behavior was also observed in NavRL and YOPO, as highlighted by the red box in Fig. \ref{fig:exp2_forest}. The limitation arises mainly from the lack of contextual memory. Without memory, the planner essentially “re-plans from scratch” at each step, which leads to decision conflicts between consecutive time steps. For example, at one stage it intends to reach the target from above. However, in the next step, the absence of memory for the previous decision causes it to replan downward. This back and forth ultimately leads to oscillatory and inefficient navigation behavior, which is precisely the NDO phenomenon we define.

Finally, regarding computational efficiency, we evaluated the inference time $T_e$ of each method. As shown in Table \ref{table:comparison_gazebo}, the traditional FAR Planner exhibits a significant increase in computational cost as the environment complexity grows. The inference time increases from 0.002$s$ in the Simple scenario to 0.120$s$ in the Forest scenario, as the enlarged environment introduces more visible nodes, resulting in higher traversal and optimization overhead during visibility graph construction. 
In comparison to traditional planners, learning-based baselines exhibit a distinct characteristic. Since these baselines directly process high-dimensional RGB-D observations, their inference time remains relatively constant and independent of the spatial scale of the environment. In contrast, CORE planner utilizes a topological graph as input. Although the scale of this graph naturally expands within larger environments like the Forest scenario, the method maintains consistently low inference latency. This efficiency is attributed to the lightweight nature of our abstract map representation. Notably, in structured (Indoor) or smaller-scale (Simple) environments, CORE planner demonstrates even faster inference speeds compared to RGB-D based baselines, as processing a sparse graph structure requires significantly less computation than handling image data.
Additionally, as the environment grows and the number of visible nodes increases, the inference time in CORE planner does increase, but remains well within acceptable real-time limits. This is confirmed by our experiments, where in the Indoor environment with approximately 750 visible nodes, the inference time is around 5 ms, and in the Forest environment with approximately 1500 visible nodes, the inference time is about 10 ms. These results further demonstrate that the proposed method scales efficiently and maintains low latency even in complex, large-scale environments.

\begin{table}[!ht]
    \centering
    \begin{threeparttable} 
    \caption{Performance comparison with SOTA methods in the Gazebo environment}
    \label{table:comparison_gazebo}
    \begin{tabular}{cccccccc}
    \toprule
    & Method & $D(m)$ & I &$T_e(s)$ & Venue(Year)\\  
    \midrule
    \multicolumn{1}{c}{\multirow{6}{*}{Simple}}  
    & CTSAC        & 24.1   &  0 &  0.006 & ICRA'25 \\ 
    & NavRL        & 25.7   &  0 &  0.010 & RAL'25  \\ 
    & YOPO         & 23.1   &  0 &  0.006 & RAL'24  \\ 
    & FAR          & 19.9   &  0 &  \textbf{0.002} & IROS'22 \\
    & CORE(ours)   & \textbf{19.7} & \textbf{0}&  0.004 &N/A  \\ 
    \midrule
    \multicolumn{1}{c}{\multirow{6}{*}{Indoor}}  
    & CTSAC        & -- \tnote{1}  &  $\infty$ \tnote{2} & 0.006 & ICRA'25\\ 
    & NavRL        & -- \tnote{1}  &  $\infty$ \tnote{2} & 0.012 & RAL'25\\ 
    & YOPO         & 1082.5        &  7                  & 0.006 & RAL'24\\ 
    & FAR          & 785.6         &  1                  & 0.015 & IROS'22\\
    & CORE(ours)   & \textbf{681.1} &\textbf{0}  & \textbf{0.005} & N/A\\ 
    \midrule
    \multicolumn{1}{c}{\multirow{6}{*}{Forest}} 
    & CTSAC        & -- \tnote{1}   &  $\infty$ \tnote{2} & \textbf{0.007} & ICRA'25\\ 
    & NavRL        & 524.2   &  2  & 0.011 & RAL'25\\ 
    & YOPO         & 389     &  0  & \textbf{0.007} & RAL'24\\ 
    & FAR          & 285.1   &  0  & 0.120 & IROS'22\\
    & CORE(ours)   & \textbf{272.0} &\textbf{0} & 0.009 & N/A\\ 
    \bottomrule
    \end{tabular}
    \begin{tablenotes}
        \item[1] Navigation failed, destination not reached.
        \item[2] Navigation failed, exceeded intervention limit.
    \end{tablenotes}
    \end{threeparttable} 
    \vspace{-12pt}
\end{table}

\vspace{-12pt}
\subsection{Ablation Study}

To analyze each component, we compare CORE with CORE w/o Sparse, which retains memory without graph sparsification, and CORE w/o Mem, which retains sparsification without memory, in Gazebo environments.

The results across the simple indoor, complex indoor, and forest environments are summarized in Table \ref{table:ablation}. Several important observations can be made.

First, the effect of graph sparsification is mainly reflected in larger-scale environments. In the Simple scenario, the identical planning time of $0.019\,\mathrm{s}$ for all three variants is expected, since the visibility graph is compact, graph sparsification is rarely activated, and the overhead of the lightweight node-wise contextual memory becomes negligible after averaging over multiple runs and rounding to the reported precision. In contrast, removing graph sparsification leads to a noticeable increase in computational overhead in large-scale environments, as reflected by the slightly higher planning time in the indoor and forest scenarios. Although disabling sparsification yields a modest improvement in travel distance, we argue that this small reduction is not as critical as achieving faster responsiveness. Moreover, the results confirm that the sparsification module effectively reduces graph complexity while preserving essential structural information.

Second, CORE w/o Mem exhibits severe oscillatory behavior and frequently falls into local optima (i.e., the NDO phenomenon we defined) due to the lack of a contextual memory mechanism. This leads to a significant increase in travel distance in complex scenarios. Specifically, compared to the full CORE planner, CORE w/o Mem requires five human interventions and incurs a 52\% increase in path length in the indoor environment, while its travel distance in the forest environment increases by 42.2\%.

Finally, the full CORE planner offers the best overall performance, consistently achieving zero human interventions, fast planning, and no oscillatory behavior. These results highlight the complementary roles of both contextual memory and graph sparsification in enabling efficient and robust navigation.

\begin{table}[!ht]
    \centering
    \begin{threeparttable} 
    \caption{Ablation study of contextual memory and graph sparsification in Gazebo environments}
    \label{table:ablation}
    \begin{tabular}{ccccccc}
    \toprule
    & Method &$D(m)$ &I &$T(s)$ & \\  
    \midrule
    \multicolumn{1}{c}{\multirow{3}{*}{Simple}}  
    & CORE w/o Sparse   & 19.7          &  0        &  0.019 \\
    & CORE w/o Mem      & 20.0          &  0        &  0.019  \\
    & CORE              & \textbf{19.7} & \textbf{0}&  \textbf{0.019} \\ 
    \midrule
    \multicolumn{1}{c}{\multirow{3}{*}{Indoor}}  
    & CORE w/o Sparse   & \textbf{670}          &  0  &  0.22 \\
    & CORE w/o Mem      & 1036.1        &  5    &  0.22\\
    & CORE              & 681.1 & \textbf{0}    &  \textbf{0.20} \\ 
    \midrule
    \multicolumn{1}{c}{\multirow{3}{*}{Forest}} 
    & CORE w/o Sparse   & \textbf{270} &  0        &  0.545 \\
    & CORE w/o Mem      & 386.8         &  0         &  0.310\\
    & CORE              & 272.0 & \textbf{0}&  \textbf{0.309} \\ 
    \bottomrule
    \end{tabular}
    \end{threeparttable}
    \vspace{-12pt}
\end{table}

\subsection{Comparison in Real-world Environments}\label{exp3}

\begin{figure}[!h]
    \centering
    \includegraphics[width=1\linewidth]{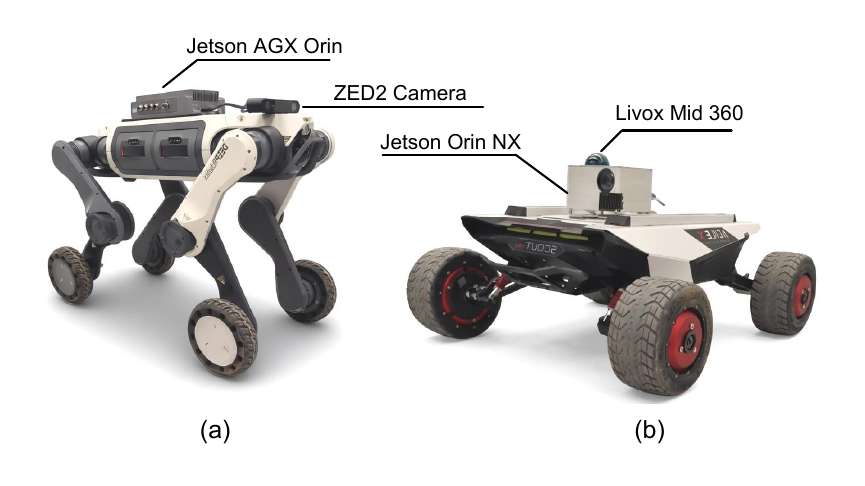}
    \caption{Experimental Robotic Platforms: LYNX M20 (a) Equipped with NVIDIA Jetson AGX Orin and ZED2 Camera, and Scout Mini (b) Equipped with NVIDIA Jetson Orin NX and Livox Mid-360 LiDAR.
}
    \label{fig:dog_car}
    \vspace{-12pt}
\end{figure}

\textbf{Experimental platforms.} Real-world validation was conducted on two heterogeneous robotic platforms to evaluate the sim-to-real zero-shot transfer capability across different sensing modalities, robot morphologies, and onboard computation units. The experimental platforms include a LYNX M20 quadruped robot (Fig.~\ref{fig:dog_car}a) equipped with a ZED2 RGB-D camera and an NVIDIA Jetson AGX Orin, and a Scout Mini unmanned ground vehicle (Fig.~\ref{fig:dog_car}b) equipped with a Livox Mid-360 LiDAR and an NVIDIA Jetson Orin NX.

\textbf{Experimental scenarios and protocol.}
Two complementary scenarios were designed to assess long-horizon planning performance and dynamic response capabilities.

\begin{figure}[!h]
    \centering
    \subfloat[]{
        \includegraphics[width=0.9\linewidth]{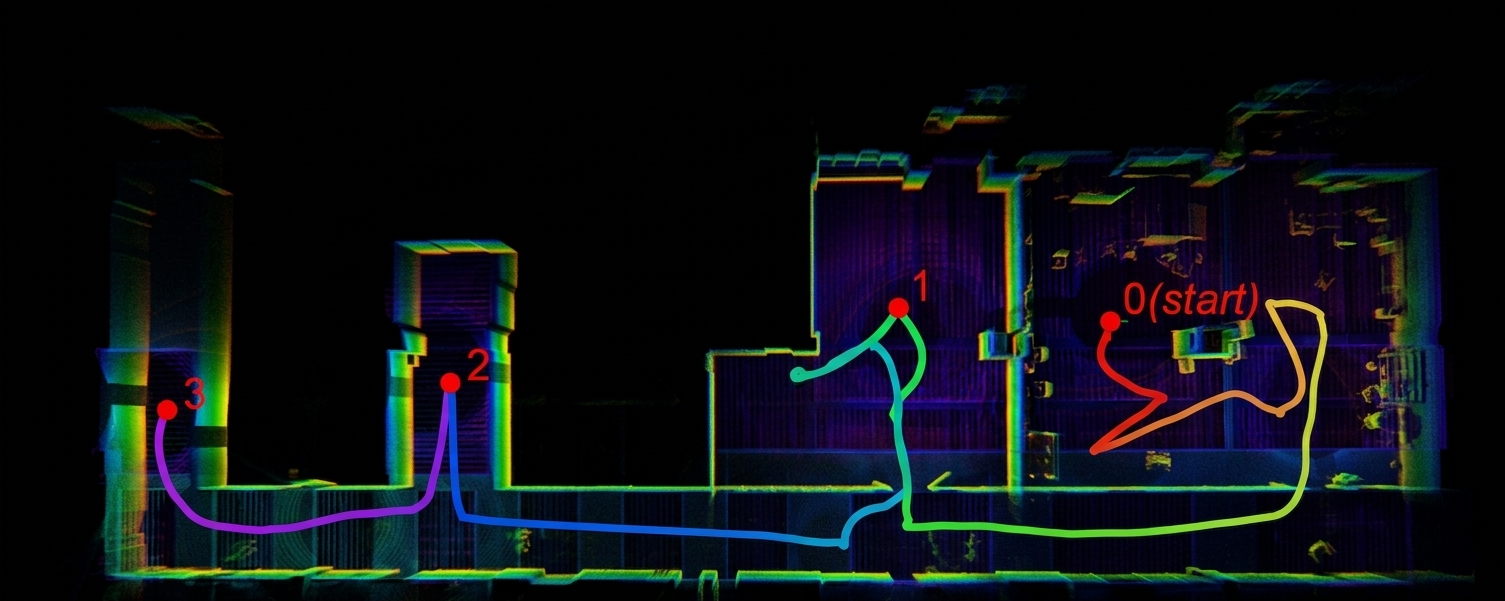}
        \label{fig:exp3_core}
    }

    \subfloat[]{
        \includegraphics[width=0.9\linewidth]{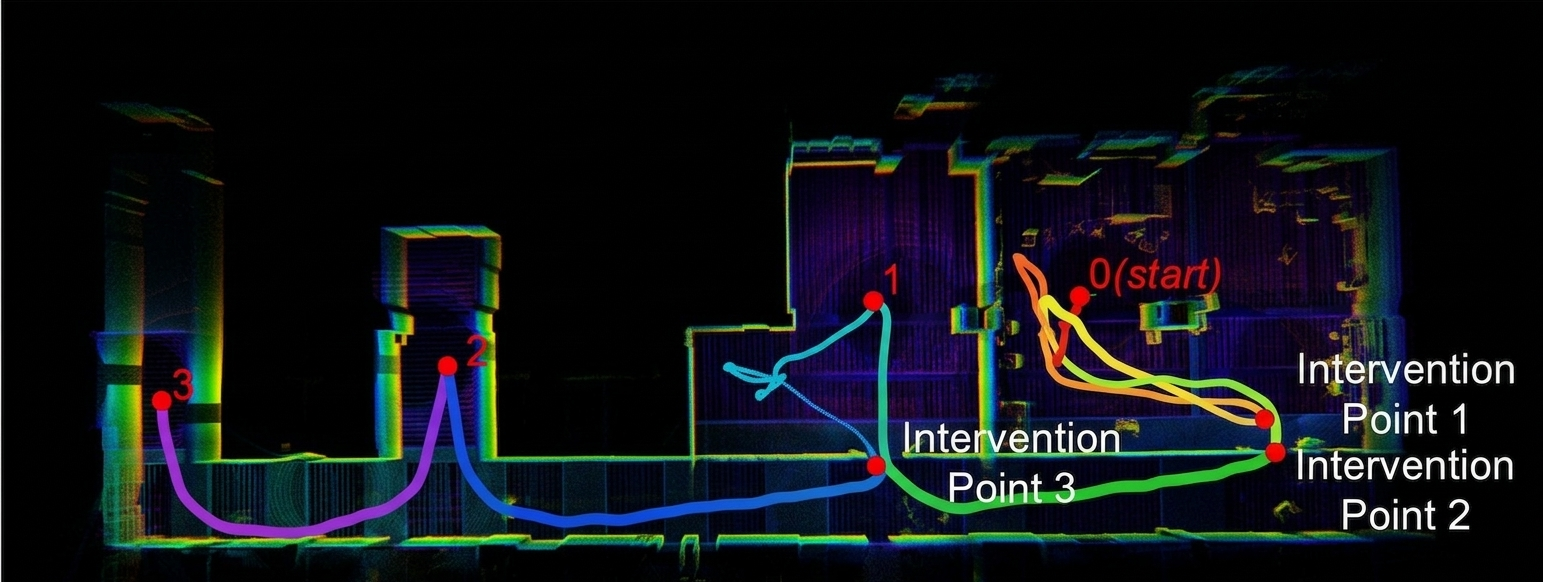}
        \label{fig:exp3_far}
    }
    \caption{
    Real-world navigation performance in a multi-room environment: (a) the proposed CORE planner successfully escapes the local optimum and completes the task autonomously, while (b) the baseline FAR Planner becomes trapped in local optima and requires repeated human intervention.
    }
    \label{fig:exp3}
    \vspace{-10pt}
\end{figure}

First, Multi-room local-optimum escape (Scout Mini). This experiment corresponds to the multi-room environment shown in Fig.~\ref{fig:exp3}. 
    The robot was required to visit multiple waypoints sequentially, with the first waypoint intentionally placed behind a nearby wall to induce a local optimum. 
    The trial was performed using the CORE planner and the FAR planner for comparison. 
    Metrics recorded include travel distance $D(m)$, planning time per decision $T(s)$, and required human interventions $I$.

    Second, Dynamic obstacle avoidance (LYNX M20). The quadruped robot was commanded to navigate through a corridor with multiple moving pedestrians. 
    The purpose of this experiment was to evaluate real-time responsiveness, collision avoidance capability, and cross-sensor generalization, since the platform relies solely on RGB-D perception instead of LiDAR.
    Depth images from the ZED2 were converted to point clouds and processed with the same visibility graph pipeline used for LiDAR input.

\textbf{Results.}
Performance of the Scout Mini is summarized in Table~\ref{table:comparison_exp3}. 
The CORE planner successfully escaped the local optimum and completed the task without any human intervention, whereas the FAR planner required multiple interventions to exit the same trap. 
Additionally, CORE achieved shorter travel distance and lower planning time per decision.

\begin{table}[!h]
    \centering
    \caption{Performance Comparison Between the CORE Planner and FAR Planner in Real-World Environments}
    \label{table:comparison_exp3}
    \begin{tabular}{cccc}
    \toprule
    Method &  $D(m)$ & $T(s)$ & I\\ 
    \midrule 
    FAR        & 132.21&0.73 & 3\\ 
    CORE(ours)   & \textbf{109.07}& \textbf{0.12} & \textbf{0}\\ 
    \bottomrule
    \end{tabular}
    \vspace{-10pt}
\end{table}

\begin{figure}[!h]
    \centering
    \includegraphics[width=1\linewidth]{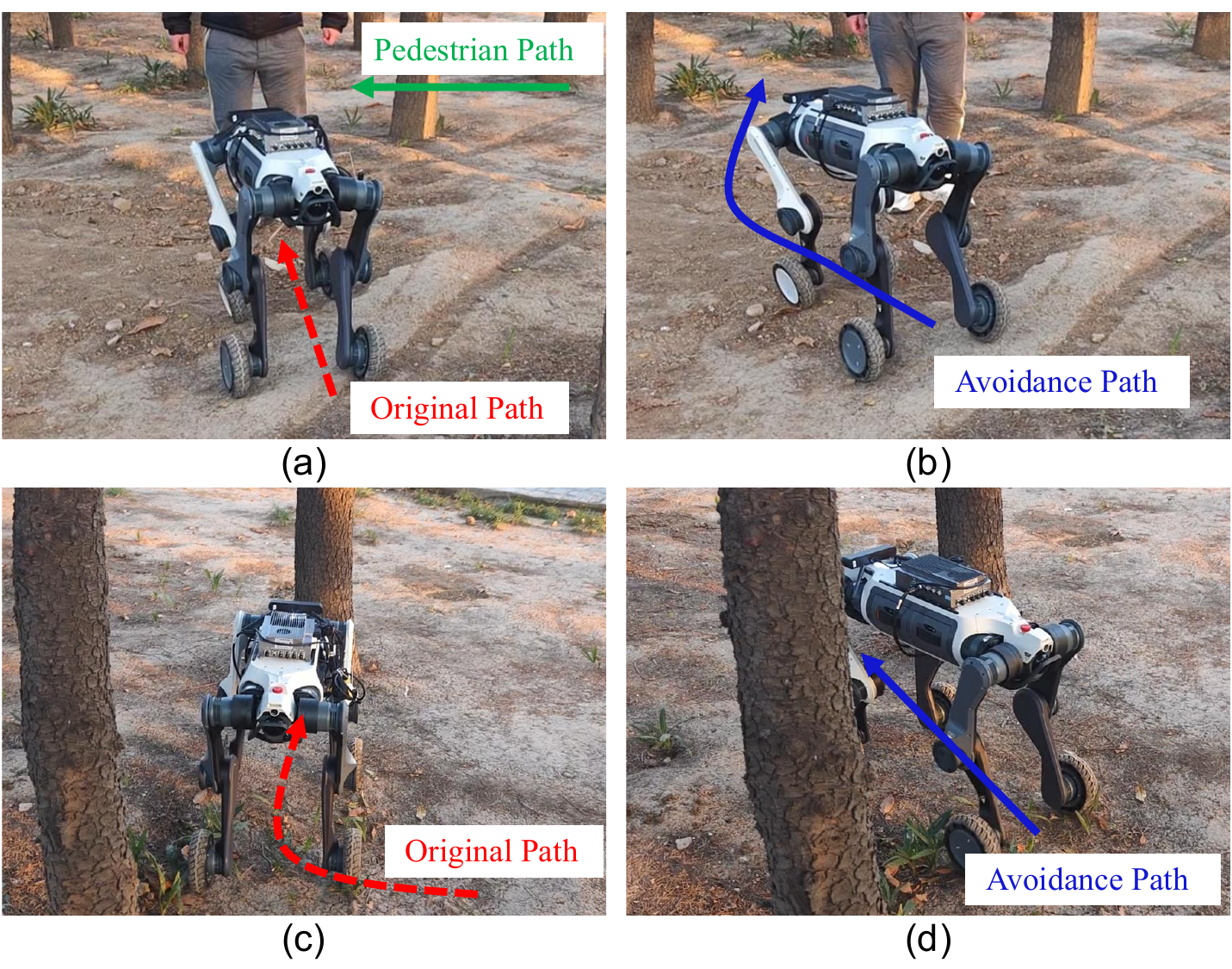}
    \caption{Illustration of the LYNX M20 Quadruped Robot Navigating Through Static and Dynamic Obstacles in a Forest Setting.
}
    \label{fig:exp4}
    \vspace{-12pt}
\end{figure}

The results of the LYNX M20 experiments are presented in Fig.~\ref{fig:exp4}. 
Despite relying on RGB-D sensing and operating on a legged platform with significantly different dynamics, CORE completed all trials without human intervention. 
The robot maintained continuous progress toward the goal, responded promptly to dynamic obstacles, and executed stable avoidance maneuvers throughout each trial. 
Representative trajectories and video demonstrations for both platforms are included in the video appendix.

\textbf{Discussion.}
The two sets of experiments confirm that the proposed planner generalizes effectively across:
(i) sensing modalities (LiDAR vs. RGB-D), 
(ii) robotic morphologies (wheeled vs. legged), and 
(iii) onboard hardware (Jetson Orin NX vs. Jetson AGX Orin).  
The planner maintained stable navigation behavior under both static and dynamic disturbances, substantiating the claimed cross-hardware and cross-sensor zero-shot sim-to-real transfer capability. The proposed CORE planner achieves the shortest navigation distance and the lowest path planning time, all without requiring any human intervention, demonstrating SOTA performance.
    
\vspace{-8pt}
\section{Conclusion} \label{conclusion}

In this work, we presented CORE planner, a novel navigation algorithm for unknown environments that integrates visibility graphs with a contextual memory mechanism. This mechanism prevents local optima entrapment and mitigates Navigation Deadlock Oscillation (NDO). Extensive experiments demonstrate that CORE achieves a 100\% success rate in image-based evaluation and reduces travel distance by up to 20\% compared with CADRL. In Gazebo simulations, CORE outperforms both classical and learning-based baselines, achieving distance reductions of 13\% over FAR Planner and 18.3–48.0\% over the learning-based methods. In real-world experiments, conducted across multiple robotic platforms and heterogeneous sensors (LiDAR and RGB-D), CORE further shortens the trajectory by 17.5\% relative to FAR Planner while requiring zero human interventions, demonstrating robustness, strong generalization across hardware configurations, and reliable zero-shot sim-to-real transfer capability. Future work will explore integrating semantic information with the current navigation framework to further enhance the system’s intelligence and adaptability.

\vspace{-10pt}
\section{Data Availability Statement}
 To promote reproducibility and facilitate future research, we have released the training code and a well-organized partial version of the codebase at \url{https://github.com/BBD00/core_planner}. The complete implementation, together with the trained models, will be made publicly available upon paper acceptance.

\begin{IEEEbiography}[{\includegraphics[width=1in,height=1.25in,clip,keepaspectratio]{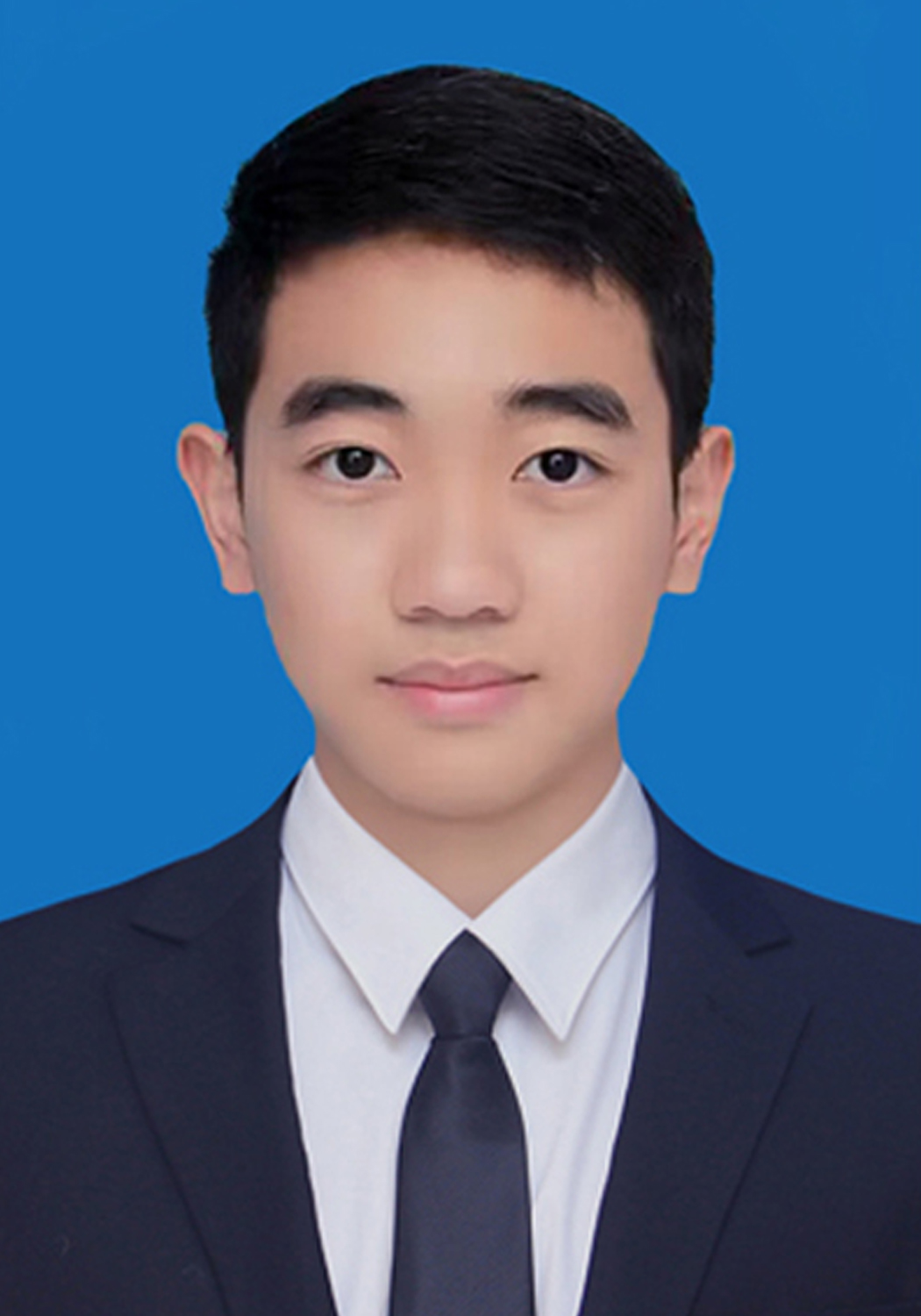}}]{Jintao Kong}
received the B.E. degree in Control Science and Engineering from the College of Automation Engineering, Nanjing University of Aeronautics and Astronautics, Nanjing, China, in 2024.
He is currently a master student in the College of Artificial Intelligence, Xi'an Jiaotong University. His research interests include path planning and autonomous driving.
\end{IEEEbiography}
\vspace{-15pt}

\begin{IEEEbiography}[{\includegraphics[width=1in,height=1.25in,clip,keepaspectratio]{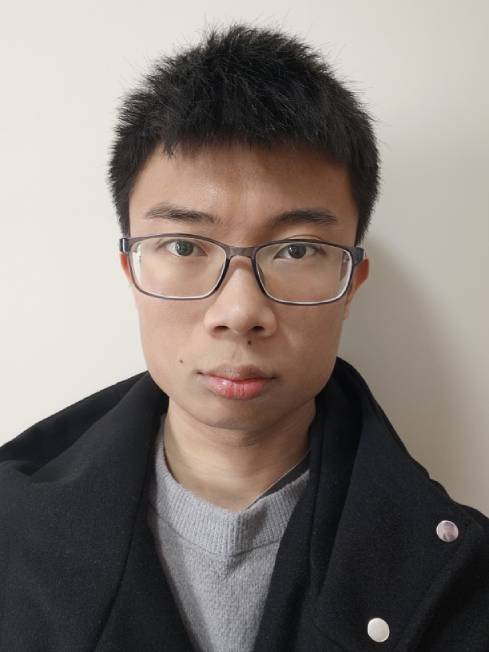}}]{Zhihao Zhang}
received the B.S. degree in Electronic Information Science and Technology from the College of Electronic Science and Engineering, Jilin University, Changchun, China, in 2024. 

He is currently a master student in the College of Artificial Intelligence, Xi'an Jiaotong University. His research interests include Computer Vision, sensor fusion,  SLAM.
\end{IEEEbiography}
\vspace{-15pt}

\begin{IEEEbiography}[{\includegraphics[width=1in,height=1.25in,clip]{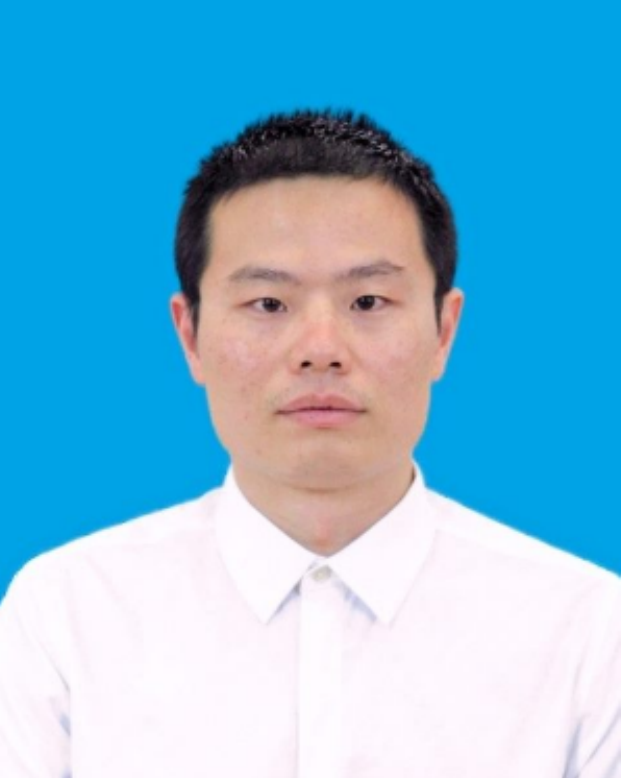}}] 
{Weihuang Chen} received the B.S. degree in thermal energy and power engineering from Harbin Engineering University in 2014, and the M.S. degree from Xi'an Jiaotong University in 2018, and the Ph.D. degree in control science and engineering from Xi'an Jiaotong University in 2023.

He is currently an assistant professor in the College of Artificial Intelligence, Xi'an Jiaotong University. His research interests include embodied intelligence robot navigation and manipulation.
\end{IEEEbiography}       
\vspace{-15pt}

\begin{IEEEbiography}[{\includegraphics[width=1in,height=1.25in,clip]{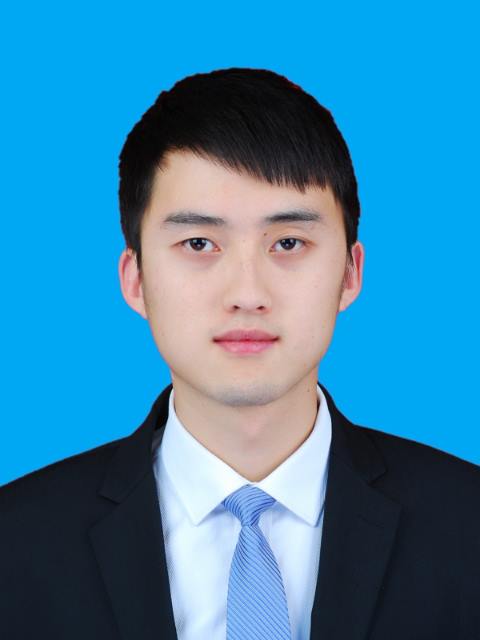}}]
{Liming Chen} received the B.S. degree and M.S. degree in mechatronic engineering from the Xi'an University of Science and Technology, Xi'an, China, in 2018 and 2021. 

He is currently a research associate in the College of Artificial Intelligence, Xi'an Jiaotong University. His research interests include state estimation, sensor fusion,  planning and control  of autonomous robots.
\end{IEEEbiography}
\vspace{-15pt}

\begin{IEEEbiography}[{\includegraphics[width=1in,height=1.25in,clip,keepaspectratio]{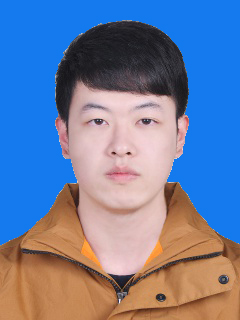}}]{Zhongyu Guo}
received the B.S. degree and M.S. degree in mechatronic engineering from the Xi'an University of Science and Technology, Xi'an, China, in 2020 and 2024. He is currently a research associate in the College of Artificial Intelligence, Xi'an Jiaotong University. His research interests include SLAM and image processing.
\end{IEEEbiography}
\vspace{-12pt}

\begin{IEEEbiography}[{\includegraphics[width=1in,height=1.25in,clip,keepaspectratio]{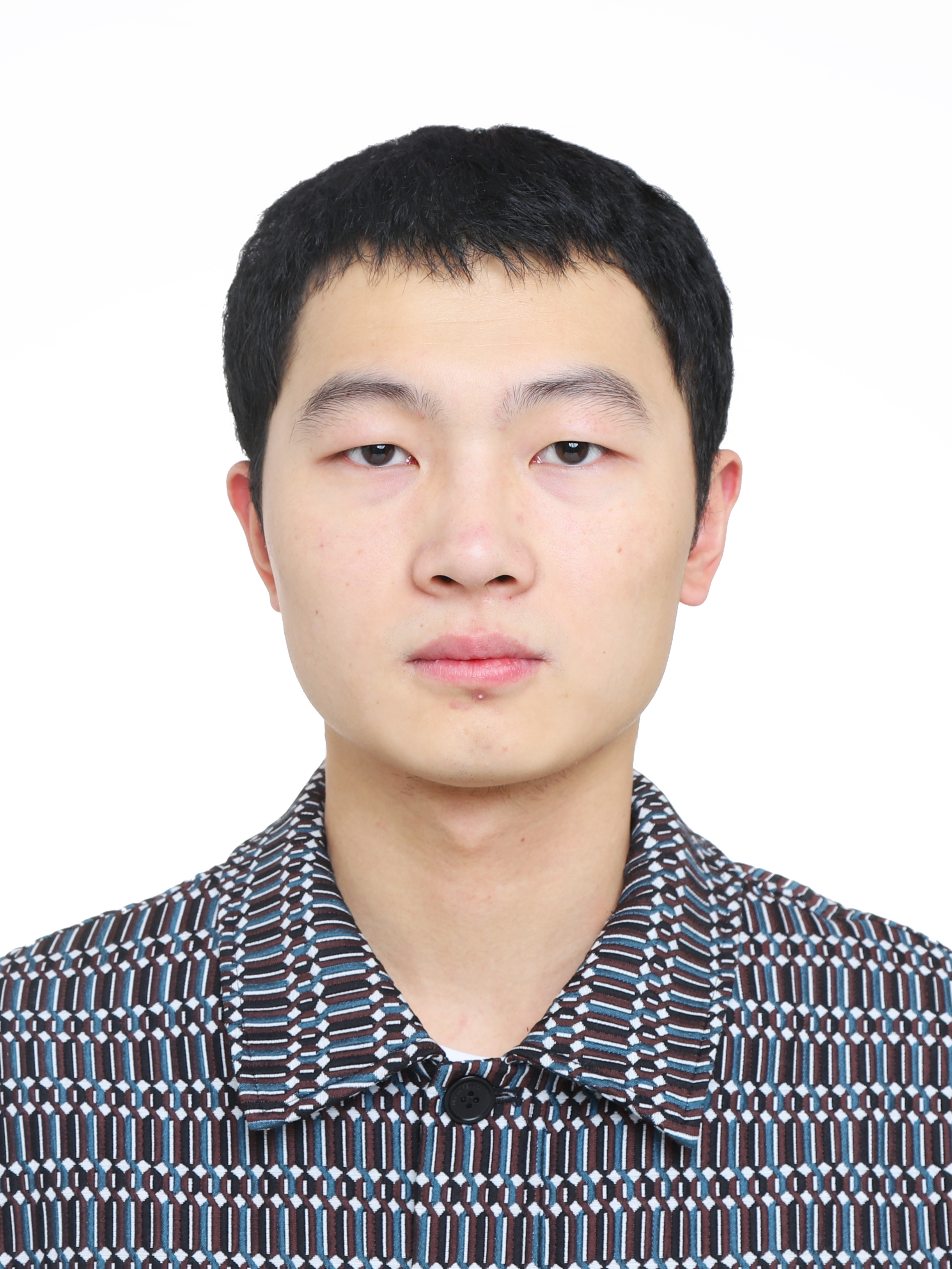}}]{Shuaiyu Liu}
received the B.S. degree in mechatronic engineering from the Beijing University of Chemical Technology, Beijing, China, in 2023. He is currently a Ph.D. candidate in the College of Artificial Intelligence at Xi'an Jiaotong University. His research interests include LLM-based RAG and VLM.
\end{IEEEbiography}

\begin{IEEEbiography}[{\includegraphics[width=1in,height=1.25in,clip]{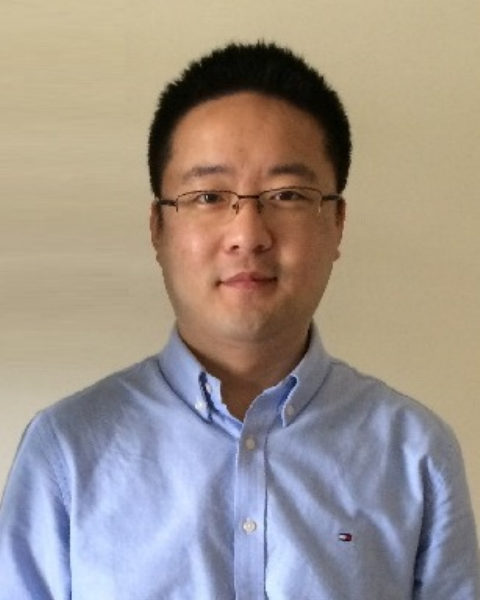}}]
{Hongbin Sun} (M'11-SM'19) received the B.S. and Ph.D. degree in electrical engineering from Xi'an Jiaotong University, Xi'an, China in 2003 and 2009, respectively. 

Currently he is a Professor of the Institute of Artificial Intelligence and Robotics, Xi'an Jiaotong University. His current research interests include autonomous vehicle, algorithm and VLSI co-design for video processing and computer vision, and hardware accelerator for deep neural network.
\end{IEEEbiography}

\vfill


\vspace{-10pt}

\begin{thebibliography}{10}
\providecommand{\url}[1]{#1}
\csname url@samestyle\endcsname
\providecommand{\newblock}{\relax}
\providecommand{\bibinfo}[2]{#2}
\providecommand{\BIBentrySTDinterwordspacing}{\spaceskip=0pt\relax}
\providecommand{\BIBentryALTinterwordstretchfactor}{4}
\providecommand{\BIBentryALTinterwordspacing}{\spaceskip=\fontdimen2\font plus
\BIBentryALTinterwordstretchfactor\fontdimen3\font minus \fontdimen4\font\relax}
\providecommand{\BIBforeignlanguage}[2]{{%
\expandafter\ifx\csname l@#1\endcsname\relax
\typeout{** WARNING: IEEEtran.bst: No hyphenation pattern has been}%
\typeout{** loaded for the language `#1'. Using the pattern for}%
\typeout{** the default language instead.}%
\else
\language=\csname l@#1\endcsname
\fi
#2}}
\providecommand{\BIBdecl}{\relax}
\BIBdecl

\bibitem{nahavandi2025comprehensive}
S.~Nahavandi, R.~Alizadehsani, D.~Nahavandi, S.~Mohamed, N.~Mohajer, M.~Rokonuzzaman, and I.~Hossain, ``A comprehensive review on autonomous navigation,'' \emph{ACM Computing Surveys}, vol.~57, no.~9, pp. 1--67, 2025.

\bibitem{al2024advancements}
S.~Al~Mahmud, A.~Kamarulariffin, A.~M. Ibrahim, and A.~J.~H. Mohideen, ``Advancements and challenges in mobile robot navigation: A comprehensive review of algorithms and potential for self-learning approaches,'' \emph{Journal of Intelligent \& Robotic Systems}, vol. 110, no.~3, p. 120, 2024.

\bibitem{wang2024online}
J.~Wang, L.~Zhao, F.~Liu, and K.~Xia, ``Online robot navigation using discrete waypoints via time-varying guidance vector fields,'' \emph{IEEE Transactions on Industrial Electronics}, 2024.

\bibitem{far}
F.~Yang, C.~Cao, H.~Zhu, J.~Oh, and J.~Zhang, ``{FAR} planner: Fast, attemptable route planner using dynamic visibility update,'' in \emph{Proc. IEEE/RSJ Int. Conf. Intell. Robots Syst. (IROS)}.\hskip 1em plus 0.5em minus 0.4em\relax IEEE, 2022, pp. 9--16.

\bibitem{navrl}
Z.~Xu, X.~Han, H.~Shen, H.~Jin, and K.~Shimada, ``{NavRL}: Learning safe flight in dynamic environments,'' \emph{IEEE Robotics and Automation Letters}, 2025.

\bibitem{yopo}
J.~Lu, X.~Zhang, H.~Shen, L.~Xu, and B.~Tian, ``You only plan once: A learning-based one-stage planner with guidance learning,'' \emph{IEEE Robotics and Automation Letters}, vol.~9, no.~7, pp. 6083--6090, 2024.

\bibitem{goal}
R.~Cimurs, I.~H. Suh, and J.~H. Lee, ``Goal-driven autonomous exploration through deep reinforcement learning,'' \emph{IEEE Robotics and Automation Letters}, vol.~7, no.~2, pp. 730--737, 2021.

\bibitem{ou2024poi}
Y.~Ou, Y.~Cai, Y.~Sun, and T.~Qin, ``Autonomous navigation by mobile robot with sensor fusion based on deep reinforcement learning,'' \emph{Sensors}, vol.~24, no.~12, p. 3895, 2024.

\bibitem{voronoi}
J.~Hu, H.~Niu, J.~Carrasco, B.~Lennox, and F.~Arvin, ``{Voronoi}-based multi-robot autonomous exploration in unknown environments via deep reinforcement learning,'' \emph{IEEE Transactions on Vehicular Technology}, vol.~69, no.~12, pp. 14\,413--14\,423, 2020.

\bibitem{huang2023goal}
W.~Huang, Y.~Zhou, X.~He, and C.~Lv, ``Goal-guided transformer-enabled reinforcement learning for efficient autonomous navigation,'' \emph{IEEE Transactions on Intelligent Transportation Systems}, vol.~25, no.~2, pp. 1832--1845, 2023.

\bibitem{attention}
A.~Vaswani, N.~Shazeer, N.~Parmar, J.~Uszkoreit, L.~Jones, A.~N. Gomez, {\L}.~Kaiser, and I.~Polosukhin, ``Attention is all you need,'' in \emph{Advances in Neural Information Processing Systems (NeurIPS)}, vol.~30, 2017.

\bibitem{brown2020language}
T.~Brown, B.~Mann, N.~Ryder, M.~Subbiah, J.~D. Kaplan, P.~Dhariwal, A.~Neelakantan, P.~Shyam, G.~Sastry, A.~Askell \emph{et~al.}, ``Language models are few-shot learners,'' in \emph{Advances in Neural Information Processing Systems (NeurIPS)}, vol.~33, 2020, pp. 1877--1901.

\bibitem{ctsac}
C.~Yang, S.~Bi, Y.~Xu, and X.~Zhang, ``{CTSAC}: Curriculum-based transformer soft actor-critic for goal-oriented robot exploration,'' \emph{arXiv preprint arXiv:2503.14254}, 2025.

\bibitem{cadrl}
J.~Liang, Z.~Wang, Y.~Cao, J.~Chiun, M.~Zhang, and G.~A. Sartoretti, ``Context-aware deep reinforcement learning for autonomous robotic navigation in unknown area,'' in \emph{Proc. Conf. Robot Learn. (CoRL)}.\hskip 1em plus 0.5em minus 0.4em\relax PMLR, 2023, pp. 1425--1436.

\bibitem{yang20253d}
Y.~Yang, H.~Yang, J.~Zhou, P.~Chen, H.~Zhang, Y.~Du, and C.~Gan, ``{3D-Mem}: {3D} scene memory for embodied exploration and reasoning,'' in \emph{Proc. IEEE/CVF Conf. Comput. Vis. Pattern Recognit. (CVPR)}, 2025, pp. 17\,294--17\,303.

\bibitem{zhu2025move}
Z.~Zhu, X.~Wang, Y.~Li, Z.~Zhang, X.~Ma, Y.~Chen, B.~Jia, W.~Liang, Q.~Yu, Z.~Deng \emph{et~al.}, ``Move to understand a {3D} scene: Bridging visual grounding and exploration for efficient and versatile embodied navigation,'' in \emph{Proc. IEEE/CVF Int. Conf. Comput. Vis. (ICCV)}, 2025, pp. 8120--8132.

\bibitem{dijkstra}
E.~W. Dijkstra, ``A note on two problems in connexion with graphs,'' in \emph{Edsger Wybe Dijkstra: His Life, Work, and Legacy}, 2022, pp. 287--290.

\bibitem{a_star}
P.~E. Hart, N.~J. Nilsson, and B.~Raphael, ``A formal basis for the heuristic determination of minimum cost paths,'' \emph{IEEE Transactions on Systems Science and Cybernetics}, vol.~4, no.~2, pp. 100--107, 1968.

\bibitem{D_star}
A.~Stentz, ``Optimal and efficient path planning for partially-known environments,'' in \emph{Proc. IEEE Int. Conf. Robot. Autom. (ICRA)}.\hskip 1em plus 0.5em minus 0.4em\relax IEEE, 1994, pp. 3310--3317.

\bibitem{lpa_star}
S.~Koenig, M.~Likhachev, and D.~Furcy, ``Lifelong planning {A}*,'' \emph{Artificial Intelligence}, vol. 155, no. 1-2, pp. 93--146, 2004.

\bibitem{D_star_lite}
S.~Koenig and M.~Likhachev, ``Fast replanning for navigation in unknown terrain,'' \emph{IEEE Transactions on Robotics}, vol.~21, no.~3, pp. 354--363, 2005.

\bibitem{rrt_star}
S.~Karaman and E.~Frazzoli, ``Sampling-based algorithms for optimal motion planning,'' \emph{The International Journal of Robotics Research}, vol.~30, no.~7, pp. 846--894, 2011.

\bibitem{rrt_connect}
J.~J. Kuffner and S.~M. LaValle, ``{RRT}-connect: An efficient approach to single-query path planning,'' in \emph{Proc. IEEE Int. Conf. Robot. Autom. (ICRA)}, vol.~2.\hskip 1em plus 0.5em minus 0.4em\relax IEEE, 2000, pp. 995--1001.

\bibitem{bit_star}
J.~D. Gammell, S.~S. Srinivasa, and T.~D. Barfoot, ``Batch informed trees ({BIT}*): Sampling-based optimal planning via the heuristically guided search of implicit random geometric graphs,'' in \emph{Proc. IEEE Int. Conf. Robot. Autom. (ICRA)}.\hskip 1em plus 0.5em minus 0.4em\relax IEEE, 2015, pp. 3067--3074.

\bibitem{wang2022chase}
C.~Wang, X.~Chen, C.~Li, R.~Song, Y.~Li, and M.~Q.-H. Meng, ``Chase and track: Toward safe and smooth trajectory planning for robotic navigation in dynamic environments,'' \emph{IEEE Transactions on Industrial Electronics}, vol.~70, no.~1, pp. 604--613, 2022.

\bibitem{rrt}
S.~M. LaValle and J.~J. Kuffner, ``Rapidly-exploring random trees: Progress and prospects,'' \emph{Algorithmic and Computational Robotics}, pp. 303--307, 2001.

\bibitem{zhang2024planner}
C.~Zhang, G.~Wang, M.~Chen, Y.~Lin, K.~Li, M.~Wu, Z.~Li, and Q.~Wang, ``{E-Planner}: An efficient path planner on a visibility graph in unknown environments,'' \emph{IEEE Transactions on Instrumentation and Measurement}, 2024.

\bibitem{li2022fps}
Q.~Li, F.~Xie, J.~Zhao, B.~Xu, J.~Yang, X.~Liu, and H.~Suo, ``{FPS}: Fast path planner algorithm based on sparse visibility graph and bidirectional breadth-first search,'' \emph{Remote Sensing}, vol.~14, no.~15, p. 3720, 2022.

\bibitem{cong2023review}
S.~Cong and Y.~Zhou, ``A review of convolutional neural network architectures and their optimizations,'' \emph{Artificial Intelligence Review}, vol.~56, no.~3, pp. 1905--1969, 2023.

\bibitem{schumann2024velma}
R.~Schumann, W.~Zhu, W.~Feng, T.-J. Fu, S.~Riezler, and W.~Y. Wang, ``{Velma}: Verbalization embodiment of {LLM} agents for vision and language navigation in street view,'' in \emph{Proc. AAAI Conf. Artif. Intell. (AAAI)}, vol.~38, no.~17, 2024, pp. 18\,924--18\,933.

\bibitem{bai2023qwen}
J.~Bai, S.~Bai, Y.~Chu, Z.~Cui, K.~Dang, X.~Deng, Y.~Fan, W.~Ge, Y.~Han, F.~Huang \emph{et~al.}, ``{Qwen} technical report,'' \emph{arXiv preprint arXiv:2309.16609}, 2023.

\bibitem{ren2024explore}
A.~Z. Ren, J.~Clark, A.~Dixit, M.~Itkina, A.~Majumdar, and D.~Sadigh, ``Explore until confident: Efficient exploration for embodied question answering,'' \emph{arXiv preprint arXiv:2403.15941}, 2024.

\bibitem{wang2025expand}
N.~Wang, W.~Chen, L.~Chen, H.~Ji, Z.~Guo, X.~Zhang, and H.~Sun, ``Expand your {SCOPE}: Semantic cognition over potential-based exploration for embodied visual navigation,'' \emph{arXiv preprint arXiv:2511.08935}, 2025.

\bibitem{vinyals2015pointer}
O.~Vinyals, M.~Fortunato, and N.~Jaitly, ``Pointer networks,'' in \emph{Advances in Neural Information Processing Systems (NeurIPS)}, vol.~28, 2015.

\bibitem{cao2023ariadne}
Y.~Cao, T.~Hou, Y.~Wang, X.~Yi, and G.~Sartoretti, ``{ARiADNE}: A reinforcement learning approach using attention-based deep networks for exploration,'' in \emph{Proc. IEEE Int. Conf. Robot. Autom. (ICRA)}.\hskip 1em plus 0.5em minus 0.4em\relax IEEE, 2023, pp. 10\,219--10\,225.

\bibitem{haarnoja2018soft}
T.~Haarnoja, A.~Zhou, P.~Abbeel, and S.~Levine, ``Soft actor-critic: Off-policy maximum entropy deep reinforcement learning with a stochastic actor,'' in \emph{Proc. Int. Conf. Mach. Learn. (ICML)}.\hskip 1em plus 0.5em minus 0.4em\relax PMLR, 2018, pp. 1861--1870.

\bibitem{chen2019self}
F.~Chen, S.~Bai, T.~Shan, and B.~Englot, ``Self-learning exploration and mapping for mobile robots via deep reinforcement learning,'' in \emph{Proc. AIAA Scitech Forum}, 2019, p. 0396.

\bibitem{env}
C.~Cao, H.~Zhu, F.~Yang, Y.~Xia, H.~Choset, J.~Oh, and J.~Zhang, ``Autonomous exploration development environment and the planning algorithms,'' in \emph{Proc. IEEE Int. Conf. Robot. Autom. (ICRA)}.\hskip 1em plus 0.5em minus 0.4em\relax IEEE, 2022, pp. 8921--8928.

\end{thebibliography}
\end{document}